\UseRawInputEncoding 
\documentclass[10pt,journal,compsoc]{IEEEtran}

\ifCLASSOPTIONcompsoc
  \usepackage[nocompress]{cite}
\else
  \usepackage{cite}
\fi

\ifCLASSINFOpdf
\else
\fi
\usepackage{amsmath}
\usepackage{ntheorem}

\usepackage{amsfonts}
\usepackage{mathrsfs}
\usepackage{amssymb}
\usepackage{lettrine}
\usepackage{algorithm}
\usepackage{algorithmic}
\usepackage{graphicx}
\usepackage{array}
\usepackage{makecell}
\usepackage{cite}
\usepackage{color}
\usepackage{booktabs}
\usepackage{url}
\usepackage{multirow}
\usepackage{footmisc}
\usepackage{bbding}
\usepackage[subnum]{cases}
\usepackage[colorlinks,
            linkcolor=blue,
            anchorcolor=blue,
            citecolor=blue,
            ]{hyperref}

\def\X{{\mathbf{X}}}
\def\U{{\mathbf{U}}}
\def\V{{\mathbf{V}}}

\def\A{{\mathbf{A}}}

\def\G{{\mathbf{G}}}
\def\E{{\mathbf{E}}}

\def\S{{\mathbf{S}}}
\def\M{{\mathbf{M}}}
\def\Y{{\mathbf{Y}}}

\DeclareMathOperator*{\argminA}{arg\,min}

\theorembodyfont{\upshape}
\theoremseparator{:}
\newtheorem{theorem}{Theorem}

\begin{document}
%
\title{Exact Decomposition of Joint Low Rankness\\ and Local Smoothness Plus Sparse Matrices}
\author{Jiangjun~Peng,
	 Yao~Wang, ~\IEEEmembership{Member,~IEEE},
	 Hongying Zhang,
	 Jianjun Wang, ~\IEEEmembership{Member,~IEEE},
	 and
	 Deyu~Meng,~\IEEEmembership{Member,~IEEE}
\thanks{J.~Peng and H.~Zhang are with School of Mathematics and Statistics and Ministry of Education Key Lab of Intelligent Networks and Network Security, Xi'an Jiaotong University, Xi'an 710049, Shaan'xi, China. (email: andrew.pengjj@gmail.com, zhyemily@mail.xjtu.edu.cn).}
\thanks{Y.~Wang is with the Center for Intelligent Decision-making and Machine Learning, School of Management, Xi’an Jiaotong University, Xi'an, Shaan'xi, China. (email: yao.s.wang@gmail.com).}
\thanks{J. Wang is with the College of Artificial Intelligence, Southwest University, Chongqing, 400715, China. (email: wjj@swu.edu.cn).}
\thanks{D.~Meng  is with the School of Mathematics and Statistics and Ministry of  Education Key Lab of  Intelligent Networks and Network Security, Xi’an Jiaotong University, Xian, Shaan'xi, China and Macau Institute of Systems Engineering, Macau University of Science and Technology, Taipa, Macau, China. (email: dymeng@mail.xjtu.edu.cn).}
}

\markboth{Journal of \LaTeX\ Class Files,~Vol.~14, No.~8, May~2021}%
{Shell \MakeLowercase{\textit{et al.}}: Bare Advanced Demo of IEEEtran.cls for IEEE Computer Society Journals}

\IEEEtitleabstractindextext{%
\begin{abstract}
It is known that the decomposition in low-rank and sparse matrices (\textbf{L+S} for short) can be achieved by several Robust PCA techniques. Besides the low rankness, the local smoothness (\textbf{LSS}) is a vitally essential prior for many real-world matrix data such as hyperspectral images and surveillance videos, which makes such matrices have low-rankness and local smoothness property at the same time. This poses an interesting question: Can we make a matrix decomposition in terms of \textbf{L\&LSS +S } form exactly?  To address this issue, we propose in this paper a new  RPCA model based on three-dimensional correlated total variation regularization (3DCTV-RPCA for short) by fully exploiting and encoding the prior expression underlying such joint low-rank and local smoothness matrices. Specifically, using a modification of Golfing scheme, we prove that under some mild assumptions, the proposed 3DCTV-RPCA model can decompose  both components exactly, which should be the first theoretical guarantee among all such related methods combining low rankness and local smoothness. In addition, by utilizing Fast Fourier Transform (FFT), we propose an efficient ADMM algorithm with a solid convergence guarantee for solving the resulting optimization problem. Finally, a series of experiments on both simulations and real applications are carried out to demonstrate the general validity of the proposed 3DCTV-RPCA model.
\end{abstract}
}

\maketitle

\IEEEdisplaynontitleabstractindextext

\IEEEpeerreviewmaketitle

%
%

\section{Introduction}

In the fields of image processing, pattern recognition and
computer vision, high-dimensional structured data are frequently encountered. In such applications, it is often reasonable to assume that an observed image has an underlying low-rank prior structure, such as hyperspectral images \cite{wang2017hyperspectral, yao2019nonconvex, peng2020enhanced} and streaming videos \cite{cao2015total, cao2016total}, and so on. However, in some real situations, such observed images are corrupted by large noise or outliers. Thus, one would like to learn this intrinsic structure to recover the true underlying image. To this end,  some matrix decomposition methods, including  Robust Principal Component Analysis (RPCA) and Probabilistic Robust Matrix Factorization (PRMF), have been proposed and attracted a great deal of attention in the recent years~\cite{wang2012probabilistic,zheng2012practical,gu2017weighted,candes2011robust, zhan2015robust, chiang2018using, chiang2016robust}.

To attain a better image recovery performance, researchers often consider exploiting other image priors in addition to the low rankness and then incorporating such priors into the matrix decomposition framework. The most popularly used prior is the so-called local smoothness~\cite{Rudin1992Nonlinear,sun2008image}. Precisely, this prior refers to how similar objects/scenes (with shapes) are adjacently distributed. As a popular prior for image data, the change in pixel values between adjacent pixels of an image is always with evident continuity over the entire image. That is, there would be a large amount of image gradients with small values in statistics.

	\begin{figure}[!]
		\centering
		\includegraphics[scale=0.21]{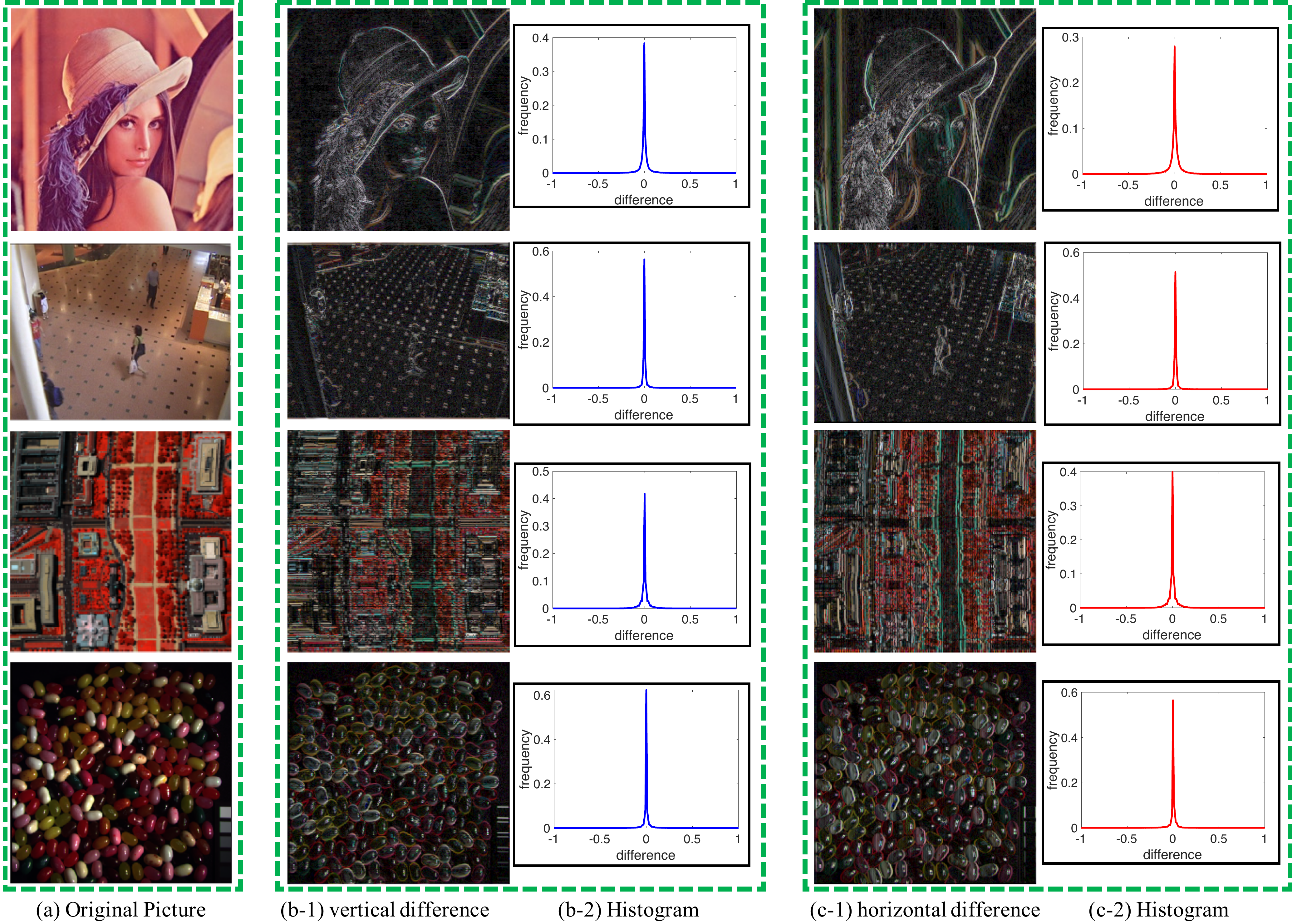}
		\vspace{-6mm}
		\caption{From top to bottom, a frame of natural, video, hyperspectral, and multispectral image (a), the gradient maps of all images in vertical dimension and their statistical histograms (b) and the gradient maps of all images in horizontal dimension and their statistical histograms (c). }
		\label{local_smoothness_overview}
		\vspace{-6mm}
	\end{figure}


By applying the first-order differential operator, we can convert the local smoothness of an image into the overall sparsity of its gradient map. This can be easily understood by observing Fig. \ref{local_smoothness_overview}, where the first column shows several images obtained from different scenes, and the second and third columns show the corresponding gradient maps and their statistical histograms, respectively. One can easily observe that although the mechanisms for generating images are different, the gradient maps all possess evident sparse configurations, which can then be naturally encoded as total variation (TV) regularization along with different spatial modes (i.e., height or width) of the image. In recent years, TV regularization has achieved much attention in natural image restoration tasks \cite{Rudin1992Nonlinear, Chambolle2004An, chambolle2010introduction, sun2008image, li1994markov}, reflecting the universality of such prior structures possessed by natural images.

It is natural for a multi-frame (or multi-spectral) image to directly impose the TV regularization on each frame (or spectrum) of such low-rank image data to deliver their local smoothness property. By using this easy manner, many efficient methods are proposed to utilize TV regularization within the low-rank decomposition framework. These methods generally take the low-rankness (\textbf{L}) and local smoothness (\textbf{LSS}) properties underlying an image as two separated parts and characterize the property of the original image by introducing two regularization terms into the model, i.e., low-rank regularization (e.g., nuclear norm) and TV regularization. Furthermore, by using $\ell_1 $ norm to characterize the usually encountered heavy-tailed and sparse noise (\textbf{S}) and integrating it into the model, many known models have been constructed, such as \cite{fan2018spatial, zeng2020hyperspectral, wang2017hyperspectral, peng2018cpct, li2018efficient, He2015Total, ji2016tensor, shi2015lrtv, li2017low, yao2019nonconvex, cao2015total}. For convenience, we call these methods as \textbf{L}\&\textbf{LSS} + \textbf{S} form.

However, these \textbf{L}\&\textbf{LSS} + \textbf{S} methods have not succeeded one important and insightful property of RPCA (with the form of $\textbf{L} + \textbf{S}$) model: There is no accurate recovery theory to guarantee to achieve an exact recovery of an original true image by using these methods. This issue of theory lacking always makes these methods lack reliability in real practice. Furthermore,  the effect of all current \textbf{L}\&\textbf{LSS} +\textbf{S}  methods highly depends on the trade-off parameters set among three items. It is always not that easy for a general user to properly pre-specify the right parameter setting to obtain a stable performance for the utilized model.

To alleviate the aforementioned issues, through sufficiently considering the intrinsic \textbf{L}\&\textbf{LSS} prior knowledge possessed by multi-frame (multi-spectral) images and building a new regularization term called correlated total variation (or CTV briefly), we propose a new RPCA model, called \textbf{CTV-RPCA}, based on this regularizer. Such a carefully constructed regularizer facilitates us to prove that the CTV-RPCA model possesses the expected exact recovery theory under some mild assumptions. That is, our model can guarantee that it can accurately separate the joint low-rank and local smoothness part as well as the sparse part with high probability from even highly polluted image data. To the best of our knowledge, this should be the first theoretical exact recovery result among all related \textbf{L}\&\textbf{LSS}+\textbf{S} studies.

Furthermore, just like RPCA \cite{candes2011robust}, based on the proposed theory, a simple choice of the trade-off parameter can be naturally suggested. Our experiments validate that such easy parameter setting always facilitates a stable performance of our CTV-PRCA method. This finely helps alleviate the parameter setting issue encountered by traditional L\&LSS +S methods along this research line.
	
In summary, this study has made the following major contributions.

1. We propose a CTV regularizer by fully considering and encoding the \textbf{L}\&\textbf{LSS} structure possessed by a general multiframe image and further present the 3DCTV-RPCA model based on three-dimensional CTV regularizer. By using a modification of Golfing scheme, the exact recovery theorem can be proved for the proposed 3DCTV-RPCA model. As far as we know, this should be the first theoretical guarantee among all related \textbf{L}\&\textbf{LSS}+\textbf{S} studies.

2. Our proposed 3DCTV-RPCA model only contains one trade-off parameter $ \lambda $ between 3DCTV regularizer and $ \ell_1 $ loss required to be tuned. Our theory naturally inspires an easy closed form for setting this parameter. Our experiments substantiate that such easy setting can help our method consistently get stable and sound performance.

3. We design an efficient algorithm by modifying the Alternating Direction Method of Multipliers (ADMM) with solid convergence guarantee and utilizing Fast Fourier Transform (FFT) to accelerate the speed to solve our proposed model.

4. Comprehensive experiments on hyperspectral image denoising, multi-spectral image denoising, and background subtraction have been implemented to validate the superiority of our method, especially its effect on finely recovering the low-rank signal from its corrupted observation for general \textbf{L}\&\textbf{LSS}+\textbf{S} scenes.

The remainder of the paper is organized as follows. In Section 2, we give the related research about the low-rank decomposition plus total variation methods. Some notations and preliminaries are provided in Section 3. In Section 4, we define the CTV regularizer and present our 3DCTV-RPCA model. Through defining gradient maps' incoherence conditions, we further prove the exact recovery theorem for the 3DCTV-RPCA model in Section 5. In Section 6, we propose the ADMM algorithm for solving our proposed model and prove its theoretical convergence. Numerical experiments are also conducted to validate the correctness of our proposed recovery theorem in this section. A series of experiments on different tasks are conducted to validate the effect and efficiency of the proposed research in Section 7. Finally, we give the conclusion and further plan.

\section{Related work}
\subsection{Robust PCA}
We first review the Robust PCA that aims to decompose an observed matrix $\M$ to obtain two of its groundtruth components, including a low-rank matrix $\X_0$ and a sparse matrix $\S_0$, that is, $ \M = \X_0 + \S_0 $.  A series of studies~\cite{wright2009robust, candes2011robust, chandrasekaran2009sparse} have shown that such a decomposition can be achieved exactly by solving the following convex objective (called Principal Component Pursuit, or PCP in brief):
\begin{equation}
\label{PCP}
\min_{\X,\S} { \| \X \|_*+ \lambda \|\S\|_1} \quad
\mbox{s.t.} \quad \M = \X + \S,
\end{equation}
where $ \|\X\|_* := \sum_{i=1}^{\scriptsize{\mbox{rank}}(\X)} \sigma_{i} $ denotes the nuclear norm of $ \X $ and  $ \|\S\|_1 := \sum_{i,j} |\S_{ij}|$ denotes the $ \ell_1 $-norm of $ \S $, and $ \lambda $ is the trade-off parameter.  More precisely, it has been proved that the model (\ref{PCP}) in RPCA \cite{candes2011robust} attains an exact separation of the underlying sparse matrix $\S_0$ and low-rank matrix $\X_0$ with high probability, if the so-called incoherence conditions are satisfied.  Along this line, a large number of works have been proposed on discussing how to embed more accurate subspace information on the low-rank matrix $\M$ into program (\ref{PCP}), e.g., \cite{2013Robust, zhan2015robust, chiang2016robust, eftekhari2018weighted, 2018Robust, sagonas2014raps, recht2010guaranteed, 2014Modified}. Among all these methods, the modified-PCP (MPCP) model\cite{zhan2015robust} and the Principal Component Pursuit with Features (PCPF) model proposed in \cite{chiang2016robust, chiang2018using}, respectively, are two of the most representative methods.
By extending the weighted $\ell_1$ norm defined by \cite{candes2008enhancing} for vectors, the weighted nuclear norm minimization (WNNM) \cite{Gu2014Weighted,gu2017weighted,xu2017multi} was proposed to help PCP program (\ref{PCP}) get better principal components. Except for considering such modifications on the low-rank matrix $ \X_0 $, there also exist some other works \cite{shang2014robust, chen2013low} that considered the weighted $\ell_1$ norm for measuring the sparse term $\S_0$.

In recent years, some works have also been proposed to extend Robust PCA to tensor cases to deal with high-dimensional data, and have been validated to be effective. If readers are interested in this issue, please read \cite{lu2016tensor, lu2019tensor, zhang2020low, bu2020hyperspectral, xue2022laplacian, xue2019nonlocal, xie2017kronecker} for more information.
\subsection{Robust PCA with Local Smoothness}
Traditional Robust PCA or low-rank decomposition framework with sparse noise form (abbreviated as $\textbf{L}+ \textbf{S}$ form) only considers the low-rankness property of data. In the following, we review the studies on combining the local smoothness property into a robust PCA framework, i.e., the studies about  $\textbf{L\&LSS}+\textbf{S}$ form.

For the local smoothness property, the total variation (TV) regularization term is generally used to characterize this prior knowledge. One can use spatial TV (STV) to embed local smoothness property among its two spatial dimensions for a single image. While for a hyperspectral image or a sequence of continuous video frames, along the third spectral or temporal dimension, some prior structures are also contained. As aforementioned, the LSS prior term is one of the most frequently utilized one. Such useful knowledge has also been extensively encoded and integrated into the recovery model. For example, similar to STV, three-dimensional total variation (3DTV), such as spectral-spatial TV (SSTV) and temporal-spatial TV (TSTV) regularizer, was proposed and widely used to characterize such local smoothness property \cite{chambolle1997image, Rudin1992Nonlinear,chambolle2004algorithm,huang2008fast}.

At present, almost all these \textbf{L}\&\textbf{LSS}+\textbf{S} methods embedding local smoothness into the robust PCA framework contain two regularization terms in one model: one is the TV regularizer representing the local smoothness (\textbf{LSS}) prior, and the other is the low-rank regularizer delivering the low rankness (\textbf{L}) prior. Based on this modeling manner, many studies have been presented by using different combinations of the TV forms (i.e., STV, SSTV, 3DTV, or other TV forms) and low-rank regularization forms (i.e., nuclear norm, low-rank decomposition, and tensor decomposition) into one model against specific tasks. Typical works include \cite{He2015Total, ji2016tensor, shi2015lrtv,li2017low,yao2019nonconvex,cao2015total, he2018hyperspectral, fan2018spatial, zeng2020hyperspectral, wang2017hyperspectral, peng2018cpct, li2018efficient, zhang2020contrast, chang2013simultaneous, chang2013robust}. There are also some recent attempts to integrate the two priors into one regularizer \cite{peng2020enhanced, li2020compressive}, which reduces the trade-off parameter to be one and makes the algorithm relatively easier to be specified.

To the best of our knowledge, for all these $\textbf{L\&LSS}+\textbf{S}$ methods, the exact decomposition theory has not been presented. The theory, however, should be critical to support the intrinsic reliability of these methods in real applications. To this aim, our main focus of this study is to propose such a useful theory to guarantee a sound implementation for such $\textbf{L\&LSS}+\textbf{S}$ model.

\section{Notations and Preliminaries}
For a given joint low rank and local smoothness tensor (e.g., a hyperspectral image) $\mathcal{X}\in\mathbb{R}^{h\times w\times s}$, where $h$, $w$ and $s$ denote the sizes of its three modes, respectively, we denote the unfolding matrix of $\mathcal{X}$ along the third mode as $\X\in\mathbb{R}^{hw\times s}$, which satisfies $\X = \mbox{unfold}(\mathcal{X})$ and $\mathcal{X} = \mbox{fold}(\X)$.
We further denote the differential operation calculated along with the $i$-th mode on $\mathcal{X}$ as $D_i$, $i = 1,2,3$, that is,
\begin{equation}\label{TensorG}
  \mathcal{G}_i = D_i(\mathcal{X}), \forall i = 1,2,3,
\end{equation}
where $\mathcal{G}_i$ $\in \mathbb{R}^{h\times w\times s}$ represents the gradient map tensor of  $\mathcal{X}$ along the $i $-th mode of the tensor\footnote{Here, we use zero paddings for $\mathcal{X}$ before applying the differential operation on it, which keeps the size of $\mathcal{G}_i$ the same as $\mathcal{X}$ and makes calculation convenient in the subsequent operations. Since the differential operation is known to be invertible, the gradient matrices $\G_i$ preserve the same rank with that of the original matrix $\X$.}. Then, we denote the unfolding matrices of these gradient maps along with spectrum mode as:
\begin{equation}\label{MatrixG}
  \G_i  = \mbox{unfold}(\mathcal{G}_i), \forall i = 1,2,3.
\end{equation}
It is easy to check that the differential operations $D_1$ and $D_2$ on $\mathcal{X}$ are equivalent to applying subtraction between rows in $\X$, and $D_3$ on $\mathcal{X}$ is equivalent to using subtraction between columns in $\mathbf{X}$, witch means that all the three different operations are linear \cite{peng2020enhanced}, the more detail about differential operations $D_i,i=1,2,3$ are introduced in supplementary material. We can then denote such three linear operations as
\begin{equation}\label{nabla_define}
  \nabla_i\X =\mbox{unfold}\left(D_i\left(\mbox{fold}(\X)\right)\right)=\G_i, \forall i = 1,2,3,
\end{equation}
where $ \nabla_i $ is defined as the corresponding differential operation on the matrix.

It should be noted that these linear operations encode the relationship between the data $\X$ and its gradient maps $\G_i$s, which help us devise an efficient method to process $\X$ by exploiting beneficial priors on such gradient maps.

\begin{figure*}[!]
		\centering
		\includegraphics[scale=0.19]{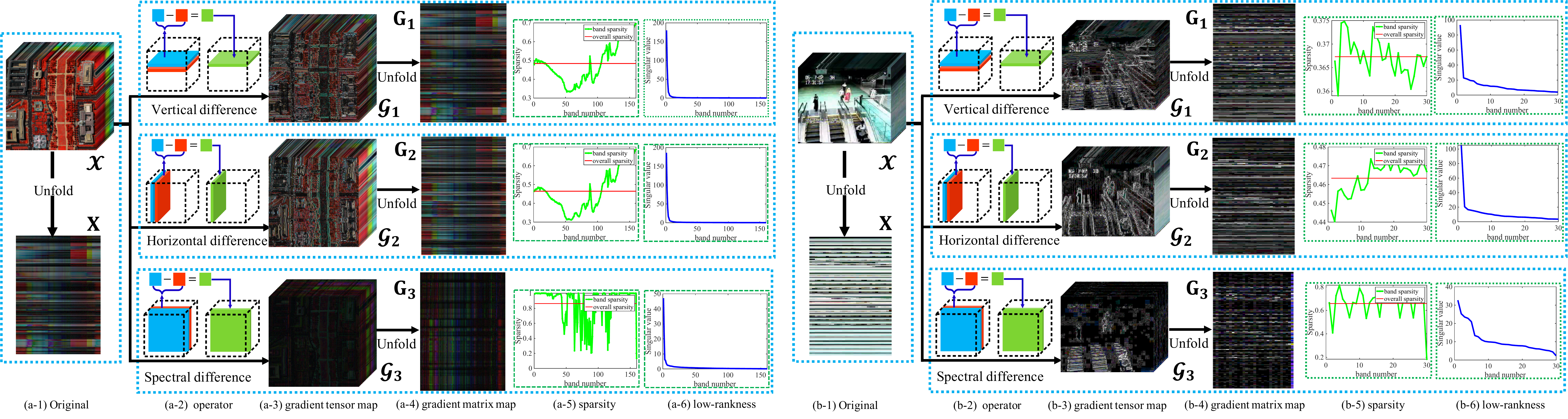}
		\vspace{-4mm}
		\caption{The illustration of correlation and sparsity of gradient maps. The left and right figures show the illustration of an HSI and streaming video data separately. The $ \mathcal{G}_i,i=1,2,3 $ is the gradient tensor map by conducting differential operator, and the $ \mathbf{G}_i,i=1,2,3 $ is the gradient matrix map by unfolding the gradient tensor map along the spectral direction. Columns (a-5) and (b-5) show the sparsity of each gradient map. Columns (a-6) and (b-6) show the singular vector curve of gradient maps. }
		\label{gradient_map_hsi}
		\vspace{-5mm}
	\end{figure*}

\section{The Correlated Total Variation Regularization}
To illustrate the main idea of the CTV regularization, we take a hyperspectral image (HSI) and a sequence of surveillance video as two illustrative examples. As shown in Fig. \ref{gradient_map_hsi}, columns (a-1), (b-1) give the ground-truths of HSI tensor/surveillance video and their unfolding matrices, and the columns (a-2)-(a-4), (b-2)-(b-4) provide the gradient maps among three dimensions.  One can easily see that all such gradient maps are sparse, i.e., the HSI image and surveillance video have evident local smoothness property. This prior of the HSI (or video) image on space and spectrum can then be naturally encoded as the TV regularization along with its different modes, that is, the $ \ell_1$-norm on the gradient maps $ \G_i (i = 1, 2, 3) $ of $ \mathcal{X}$, where $\G_1$, $\G_2$ and $\G_3$ represent unfolding matrices of the gradient maps of $\mathcal{X} $ calculated along its spatial width, height and spectrum mode, respectively. This term is called the 3DTV regularization with the form:
\begin{equation}
\label{3DTV}
\| \mathcal{X} \|_{\scriptsize{\mbox{3DTV}}} = \sum_{i=1}^3 \| \G_i\|_1 = \sum_{i=1}^3 \{ \sum_{k=1}^s \| \G_i (:,k)\|_1 \}.
\end{equation}
It has been validated that this term can be beneficial
for various multi-dimensional data processing tasks, e.g., \cite{He2015Total, Yao2018Hyperspectral, cao2016total}. 

Because the difference operator is a linear operator with approximately full rank, the gradient map obtained by imposing the difference operation on the original data also inherits the low rank of the original data, thus we have $ \mbox{rank}(\G_i)=\mbox{rank}(\X), (i=1,2,3)$. Therefore, except for sparsity, there always has a strong correlation between gradient maps, which can be seen in Fig. \ref{gradient_map_hsi}(a-6) and (b-6). In this figure, we take the SVD operator on gradient maps and observe that only a few singular values are significant, implying that the gradient maps possess low-rank structures. Furthermore, one can find that the sparsity patterns of gradient maps $ \mathcal{G}_i $s are different,  just as the Fig. \ref{gradient_map_hsi}(a-5) and (b-5) shown. This implies that one should treat the gradient maps differently. Since 3DTV mentioned above obviously treats such gradient maps equally, it might insufficiently characterize the strong correlation property and unique sparsity for each image. This motivates us to consider a more rational TV-based regularization for more faithfully and concisely encoding such prior knowledge.

To that end, we treat the gradient map of low-rank image data as a whole. As shown in Fig. \ref{gradient_map_hsi}(a-6) and (b-6), we can use the nuclear norm to describe the correlation among these gradient maps, which we call it correlated total variation (CTV) regularization. More precisely, the CTV along the $i$-th dimension is:
\begin{equation}
\label{ctv}
\| \X \|_{\scriptsize{\mbox{CTV}}} = \| \nabla_1 \X\|_* =  \| \G_i\|_*.
\end{equation}
Since we know that $ \mbox{rank}(\G_i) = \mbox{rank}(\X)$ for each gradient map, it is then evident that CTV also conveys the low-rankness of the matrix as well as the nuclear norm on $\X$ does.
Besides, according to the compatibility of matrix norms, the following inequality holds: $ \| \G_i\|_{\tiny{\mbox{F}}} \leq \| \G_i\|_* \leq \| \G_i\|_1$, where $ \| \G_i\|_{\tiny{\mbox{F}}} $ and $ \| \G_i\|_1 $ correspond to the isotropic TV regularization \cite{Rudin1992Nonlinear} and anisotropic TV regularization \cite{chambolle2010introduction, osher2005iterative}, respectively. It can then be easily seen that minimizing the CTV regularizer tends to essentially minimize the TV one. This implies that CTV is expected to encode the local smoothness prior structure possessed by underlying data. Thus, CTV is expected to be able to integrate the low rankness and local smoothness into one unique term.

%

Similar to Eq. (\ref{ctv}),  it is natural to define the following 3DCTV regularization:
\begin{equation}
\label{3dnntv}
\| \X \|_{\scriptsize{\mbox{3DCTV}}} = \sum_{i=1}^3 \| \nabla_i \X\|_* = \sum_{i=1}^3 \| \G_i\|_*.
\end{equation}

Similar as the previous L\& LSS + S models,  we can then employ the above CTV regularizer to attain the following ameliorated RPCA optimization problem:
\begin{equation}
\label{NNTV}
\begin{split}
\min_{\X,\S} \ \  &  \| \X \|_{\scriptsize{\mbox{3DCTV}}}  + 3 \lambda \| \S \|_1 \\
\mbox{s.t.} \ \ &  \M = \X + \S,
\end{split}
\end{equation}
where $ \lambda $ is the trade-off parameter to balance 3DCTV regularization and $ \ell_1 $-norm. It is easy to see that the above model is a RPCA-type model, and thus we call it 3DCTV-RPCA throughout this paper. Furthermore, substituting Eq. (\ref{nabla_define}) and Eq. (\ref{3dnntv}) into Eq. (\ref{NNTV}), we get that Eq. (\ref{NNTV}) can be equivalently expressed as
\begin{equation}
\label{3dnntv_r}
\begin{split}
\min_{\X,\S} \quad& \sum_{i=1}^{3}{\|\mathbf{G}_i\|_*} + 3\lambda \|\S\|_1\\
\mbox{s.t.} \quad & \mathbf{M} = \X +\S, \\
& \G_i = \nabla_{i}(\X) ,i=1,2,3.
\end{split}
\end{equation}

Next we will give the exact recovery theorem that asserts that under some weak conditions, 3DCTV-RPCA model (\ref{3dnntv_r}) can accurately separate the joint low-rank and local smoothness part $ \X $ and the sparse part $ \S$ with high probability.

\section{Recovery Guarantee of 3DCTV-RPCA}
For all the RPCA-type models, the incoherence condition is a vital assumption on low rank component. Unlike the PCP model (\ref{PCP}) proposed in RPCA\cite{candes2011robust}, 3DCTV-RPCA model (\ref{3dnntv_r}) is proposed to separate the joint low rank and local smoothness component $ \X_0 $ and sparse component $ \S_0 $ from the contaminated matrix $ \M = \X_0 + \S_0 $. We have analyzed in Section 4 that the nuclear norm on the gradient map (i.e., CTV norm) can simultaneously encode low-rank and local smoothness properties. Therefore, we assume the incoherence condition on the gradient map (i.e., $ \G_i,i=1,2,3 $) instead of the original matrix $ \X_0 $.
\subsection{Incoherence Conditions}
The incoherence condition is proposed to constrain that the left and right singular value vectors of the recovered low-rank components should not be highly concentrated \cite{candes2011robust, zhan2015robust}. With this understanding, many studies on incoherence conditions have been put forward, such as \cite{chen2015incoherence, zhan2015robust, 2011Recovering}. These assumptions about incoherence conditions can all be applied to our 3DCTV-RPCA model (\ref{3dnntv_r}). For the convenience, we choose the incoherence conditions defined in RPCA \cite{candes2011robust} to define our incoherence conditions on gradient maps.

Suppose that each gradient map $ \G_i (i=1,2,3) $ of $ \X_0 $ has the singular value decomposition $ \U_i \Sigma_i \V_i^T $, where $ \U_i \in \mathbb{R}^{n_1\times r}, \V_i \in \mathbb{R}^{n_2 \times r}$, and then the incoherence conditions hold with a constant $ \mu $ for (\ref{3dnntv_r}) are assumed as:
\begin{equation}
\label{new_U}
 \max_k   \|\U_i^T \hat{e}_k\|^2 \leq \frac{\mu r}{n_1}, i=1,2,3,
\end{equation}
\begin{equation}
\label{new_V}
 \max_k   \|\V_{i}^T \hat{e}_k\|^2 \leq \frac{\mu r}{n_2}, i=1,2,3,
\end{equation}
and
\begin{equation}
\label{new_UV}
\|\U_{i}\V_{i}^T\|_\infty \leq \sqrt{\frac{\mu r}{n_1 n_2}}, i=1,2,3,
\end{equation}
where $ \hat{e}_k$ is the standard orthonormal basis, and  $ \| \M \|_{\infty} = \max_{i,j} |\M_{i,j}|$. The first two incoherence conditions (\ref{new_U}) and (\ref{new_V}) imply that the matrix $ \U $ and $ \V $ should not be highly concentrated on the matrix basis. The third condition (\ref{new_UV}) constrains the inner product between the rows of two basis matrices. It is not hard to check that, by using Cauchy-Schwartz inequality, the first two conditions only imply that $ \|\U\V^T\|_\infty \leq \|\U^T \hat{e_k}\| \|\V^T \hat{e_k}\| \leq \frac{\mu r}{\sqrt{n_1 n_2}} = \frac{\sqrt{\mu r}}{\sqrt{n_1 n_2}} \sqrt{\mu r} $, which is looser than what the third condition requires. Therefore, the third constraint is also called the joint (or strong) incoherence condition, which is unavoidable for the PCP program~\cite{chen2015incoherence}.

\subsection{Main Theorem}
Based on the incoherence conditions (\ref{new_U})-(\ref{new_UV}), we can get the following exact decomposition theorem.
\begin{theorem}
\label{theorem_main2} \textbf{\emph{Suppose that the each gradient map $\G_i,i=1,2,3$ of joint low rank and local smoothness matrix $ \X_0 \in \mathbb{R}^{n_1\times n_2}$ obey the incoherence conditions (\ref{new_U})-(\ref{new_UV}), and the support set $ \Omega $ of $ \S_0 $ is uniformly distributed among all sets of cardinality $ m $. Then there is a numerical constant $ c $ such that with probability at least $ 1-cn_{(1)}^{-10} $ (over the choice of support of $ \S_0 $), 3DCTV-RPCA model (\ref{3dnntv_r}) with $ \lambda = 1/\sqrt{n_{(1)}}$ is exact, i.e. $ \hat{\X}=\X_0$ and $\hat{\S}=\S_0 $, provided that
\begin{equation}
\label{pho_r}
\mbox{rank}(\X_0) \leq \rho_r n_{2} \mu^{-1}(\log n_{(1)})^{-2} ~ \mbox{and} ~ m \leq \rho_s n_1n_2,
\end{equation}
where $ \rho_r $ and $ \rho_s $ are positive numerical constants. $ \rho_s $ determines the sparsity of $ \S_0 $, and $ \rho_r $ is a small constant related to the rank of $ \X_0 $.}}
\end{theorem}

The proof of the above theorem is placed in the supplementary material. The above theorem asserts that when $ \G_i,i=1,2,3$ satisfies the incoherence conditions (\ref{new_U})-(\ref{new_UV}), and the location distribution of $ \S_0 $ satisfies the assumption of randomness, the 3DCTV-RPCA model (\ref{3dnntv_r}) is able to exactly recover the joint low-rank and local smoothness component and the sparse component with high probability. Specifically, Eq. (\ref{pho_r}) implies that the upper bound of rank of the recoverable matrix $ \X_0 $ is ​​inversely proportional to the constant $\mu $. Therefore, a smaller $\mu$ can bring a better separation effect. When incoherence conditions are applied in original matrix $ \X_0 $, Theorem \ref{theorem_main2} can also be suitable for PCP model (\ref{PCP}). Owing that the $ \G_i,i=1,2,3 $ has same rank with $ \X_0 $, and 3DCTV regularization can simultaneously encode the L\& LSS prior, thus 3DCTV-RPCA model is more suitable than PCP model for joint low rank and local smoothness data, which is reflected in the theorem that the $ \mu $ required by 3DCTV-RPCA model will be smaller from a statistical point of view.  In Section 6.3, we also provide an empirical analysis for the $ \mu $ needed for incoherence conditions (\ref{new_U})-(\ref{new_UV}) to validate the above assertion.

As for the choice of trade-off coefficient $ \lambda $, we expect to select the proper setting to balance the two terms in $ \| \G_i\|_* + \lambda \| \S \|_1, i=1,2,3$ appropriately based on the understanding of the data. In Theorem \ref{theorem_main2}, our theoretical assertion indicates that the parameter $ \lambda = 1/\sqrt{n_{(1)}} $ leads to an appropriate recovery for 3DCTV-RPCA model. In this sense, $\lambda = 1/\sqrt{n_{(1)}} $ is universal. More detailed discussions on this hyper-parameter setting issue is presented in the supplementary material.

\subsection{Flow of the Proof}
The proof of Theorem \ref{theorem_main2} can be divided into three parts. In the first part, we convert the gradient map $ \G_i,(i=1,2,3) $ in the 3DCTV-RPCA model (\ref{3dnntv_r}) into the product of the specific matrix and the original matrix $\X_0$, and then give the equivalent model of the 3DCTV-RPCA model to facilitate us to complete the proof of Theorem (\ref{theorem_main2}). In the second part, we introduce the elimination lemma, which proves that if the 3DCTV-RPCA model (\ref{3dnntv_r}) can exactly separate sparse component, and low-rank and local smoothness components from $ \M_0=\X_0 + \S_0 $, then by reducing the sparsity of $ \S_0 $, both components can also be accurately separated. Finally, in the third part, we introduce the dual certification to finish the proof. Specifically, we provide a way for the construction of the solution to the 3DCTV-RPCA model, and then prove that the constructed solution is exact to the 3DCTV-RPCA model (\ref{3dnntv_r}).

It should be noted that we need to construct the dual certification about the gradient map $ \G_i,(i=1,2,3) $ of original matrix rather than the original matrix $ \X_0 $. And in the proof of dual certification, we introduce two key inequalities about matrix norm to make such certification be satisfied. More details about the proof of Theorem \ref{theorem_main2} are presented in the supplementary material due to the page limitation.
\section{ Optimization Algorithm and Numerical Experiments}
\subsection{Optimization by ADMM}
Here we use the well-known alternating direction method of multipliers (ADMM) introduced by \cite{lin2010augmented, boyd2011distributed} to derive an efficient algorithm for solving (\ref{3dnntv_r}) with convergence guarantee. Based on the ADMM methodology, we first write the augmented Lagrangian function of  Eq. (\ref{3dnntv_r}) as:

\begin{equation}
\label{ALM_solution}
\begin{split}
\mathcal{L}(\mathbf{X},\S,\{ \mathbf{\Gamma}_i \}_{i=1}^4, &\{ \G_i \}_{i=1}^3) = \sum_{i=1}^{3} \| \G_i \|_* + 3\lambda \|\S\|_1 \\
  & +\sum_{i=1}^{3} \frac{\mu}{2}\|\nabla_{i}\mathbf{X} -\G_i + \frac{\mathbf{\Gamma}_{i}}{\mu}\|_F^2 \\
& + \frac{\mu}{2} \|\M-\X-\S + \frac{\mathbf{\Gamma}_{4}}{\mu} \|_F^2,
 \end{split}
\end{equation}

where $ \mu $ is the penalty parameter, and $ \mathbf{\Gamma}_{i}, (i = 1, 2, 3,4) $ are the Lagrange multipliers. We then shall discuss how to solve its sub-problems for each involved variable.

\subsubsection{Computing $\S^{k+1} $}
Fixing other variables except for $\S $ in Eq. (\ref{ALM_solution}), we can obtain the following sub-problem:
\begin{equation}
\label{solveE}
\argminA_{\S} 3 \lambda \|\S\|_1 + \frac{\mu}{2} \|\S - ( \M-\X^{k}+\frac{\mathbf{\Gamma_4}}{\mu} ) \|_F^2,
\end{equation}
and the solution of the above problem can be expressed as
$ \S^{k+1}= \mathcal{S}_{3\lambda/\mu}(\M-\X^{k}+\mathbf{\Gamma}_4^{k}/\mu) $, where $ \mathcal{S}$ is the soft-threthresholding operator defined by \cite{donoho1995noising}.

\subsubsection{Computing ($\X^{k+1} $, $\G_i^{k+1}$)}
We first update $ \G_i $ by solving the following sub-problem:
\begin{equation*}
\argminA_{\G_i} \|\G_i \|_* + \frac{\mu}{2} \| \G_i  - ( \nabla_{i} \X^{k}+ \frac{\mathbf{\Gamma}_i^{k}}{\mu}) \|_F^2.
\end{equation*}
The solution of this sub-problem is:
\begin{equation}
\label{solveG}
\left\{
\begin{split}
\G_i^{k+1} &= \mathbf{U} \mathcal{S}_{1/\mu}(\mathbf{\Sigma}) \mathbf{V}^{T}, \\
\U\mathbf{\Sigma}\V^{T}  & = \mbox{svd}(\nabla_{i}\X^{k} + \mathbf{\Gamma}_i^{k}/\mu,\mbox{'econ'}).\\
\end{split}
\right.
\end{equation}
Next we update $ \X $ by solving the following sub-problem:
\begin{equation*}
\begin{split}
\argminA_{\X} &\sum_{1}^3 \frac{\mu}{2} \|\nabla_{i}\mathbf{X}-\G_i^{k+1} + \mathbf{\Gamma}_i^{k}/\mu \|_F^2  \\
&+ \frac{\mu}{2} \|\Y-\X-\S^{k+1} + \mathbf{\Gamma}_4^{k}/\mu\|_F^2.
\end{split}
\end{equation*}
Optimizing the above problem can be treated as solving the following linear system:
\begin{equation}
\label{nabla_update}
\begin{split}
&\left(\mu\mathbf{I}+\mu\sum_{i=1}^3\mathbf{\nabla}_i^T\mathbf{\nabla}_i\right)(\mathbf{X}) =\\
& \mu(\M-\S^{k+1})+\mathbf{\Gamma}_4^{k} +\mu\sum_{i=1}^3\mathbf{\nabla}_i^T\left(\G_i^{k+1}\right)-\mathbf{\nabla}_i^{T}(\mathbf{\Gamma}_i^{k}),\\
\end{split}
\end{equation}
where  $\mathbf{\nabla}_i^{T}(\cdot)$ indicates the  transpose operator of ${\nabla}_i(\cdot)$\footnote{Since $ \mathbf{\nabla}_i(\cdot)$ is a linear operator on $\X$, there exists a matrix $\A_i$ which makes the operation $\A_i\cdot\mbox{vec}(\X)$ equivalent to $ \mathbf{\nabla}_i(\X)$. Then, $ \mathbf{\nabla}_i^{T}(\cdot)$ means the operator equivalent to the transposed matrix $\A_i^{T}$.}. Attributed to the block-circulant structure of the matrix corresponding to the operator $ \nabla_{i}^{T} \nabla_{i} $, it can be diagonalized by the 3D FFT matrix. Specifically, similar to~\cite{krishnan2009fast}, by performing Fourier transform on both sides of Eq. (\ref{nabla_update}) and adopting the convolution theorem, the closed-form solution to $\X^{k+1}$ can be easily deduced as
\begin{equation}
\label{solveX}
\left\{
\begin{split}
&{\mathbf{H}}=\sum_{n=1}^3\mathcal{F}\left(\mathbf{D}_n\right)^*\!\odot\!\mathcal{F}\left(\mbox{fold}\left(\mu\G_i^{k+1}- \mathbf{\Gamma}_i^{k} \right)\right),\\
&\mathbf{T}_x=|\mathcal{F}(\mathbf{D}_1)|^2+\mathcal{F}(\mathbf{D}_2)|^2+|\mathcal{F}(\mathbf{D}_3)|^2, \\
&\X^{k+1}=\mathcal{F}^{-1}\left(\frac{ \mathcal{F}\left(\mbox{fold}(\mu\M-\mu\S^{k+1}+\mathbf{\Gamma}_4^{k}) \right) +\mathbf{H}}{\mu{\mathbf{1}}+\mu\mathbf{T}_x}\right),
\end{split}
\right.
\end{equation}
where $\mathbf{1}$ represents the tensor with all  elements as $1$ , $\odot$ is the element-wise multiplication, $\mathcal{F}(\cdot)$ is the Fourier transform, and
$\left|\cdot\right|^2$ is the element-wise square operation.
\subsubsection{Computing Multipliers $ \mathbf{\Gamma}_i^{k+1},i=1,2,3,4 $}
Based on the general ADMM principle, the multipliers are further updated by the following equations:
\begin{equation}
\label{solveM}
\left\{
\begin{split}
\mathbf{\Gamma}_i^{k+1}&=\mathbf{\Gamma}_i^{k}+\mu\left(\nabla_i\X^{k+1}-\G_i^{k+1} \right),n=1,2,3,\\
\mathbf{\Gamma}_4^{k+1}&=\mathbf{\Gamma}_4^{k}+\mu\left(\M-\X^{k+1}-\S^{k+1} \right), \\
\mu & = \mu\rho,\ \\
\end{split}
\right.
\end{equation}
where $ \rho $ is a constant value greater than 1.

Summarizing the aforementioned descriptions, we can get the  following {Algorithm} 1.

\begin{algorithm}
\caption{ Algorithm for solving 3DCTV-RPCA model.}\label{alg_denoise}
\small
\begin{algorithmic}[1]
\renewcommand{\algorithmicrequire}{\textbf{Input:}}
\renewcommand{\algorithmicensure}{\textbf{End}}
\REQUIRE The low rank image data $\mathbf{\mathcal{M}}\in\mathbb{R}^{h\times w\times s}$, unfolding to the matrix $\mathbf{M}\in\mathbb{R}^{hw\times s}$, $ \lambda = 1/\sqrt{hw}$ and $ \epsilon_1 = \epsilon_2 = 10^{-6}$.

\renewcommand{\algorithmicrequire}{\textbf{Initialization:}}
\renewcommand{\algorithmicensure}{\textbf{End}}
\REQUIRE Initial $\X = \mbox{randn}(hw,s)$, $\S = \mathbf{0}$.
\WHILE {not converge}
\STATE
Update $\G_i^{k+1}$ by Eq. (\ref{solveG}).
\STATE
Update $\S^{k+1} $ by Eq. (\ref{solveE}).
\STATE
Update  $\X^{k+1} $ by Eq. (\ref{solveX}).
\STATE
Update $\mathbf{\Gamma}_i^{k+1}$ by Eq. (\ref{solveM}).
\STATE
$ \mu = \rho \mu, k:=k+1 $
\STATE
Check the convergence conditions\\
$\quad \| \M-\X^{k+1}-\S^{k+1} \|_{\footnotesize{\mbox{F}}}^2/\|\M\|_{\footnotesize{\mbox{F}}}^2\leq\epsilon_1$,\\
$\quad \Vert\nabla_{i}\X^{k+1}-\G_i^{k+1}\|_{\footnotesize{\mbox{F}}}^2/\| \M \|_{\footnotesize{\mbox{F}}}^2\leq\epsilon_2,n=1,2,3$.\\
\ENDWHILE
\renewcommand{\algorithmicrequire}{\textbf{Output:}}
\renewcommand{\algorithmicensure}{\textbf{End}}
\REQUIRE  $\mbox{Fold}(\X^{k+1})\in\mathbb{R}^{h\times w\times s}$.
\end{algorithmic}
\end{algorithm}

\subsection{Complexity Analysis}
\label{Running time}
Following the procedure of Algorithm \ref{alg_denoise}, the main computational complexity of each iteration includes FFT and SVD operations. Suppose that the size of the calculated data is $n_1\times n_2$ and $ n_1\geq n_2 $ without loss of generality. The complexities of FFT and SVD operations are $\mathcal{O}(n_1n_2\log (n_1))$ and $\mathcal{O}(n_1n_2^2)$, respectively. Our algorithm needs one FFT operation and three SVD operations. Thus the computation cost of Algorithm \ref{alg_denoise} is around $\mathcal{O} (n_1n_2.(\log (n_1)+3n_2))$, which is similar to the previous \textbf{L}\&\textbf{LSS} method, such as LRTV \cite{He2015Total}, which needs one FFT operation and one SVD operation. In Table \ref{table:complexity}, we present the prior characterizations and the computational complexities of some models for easy comparison, and we use L, LSS-S, and LSS-ST in Table \ref{table:complexity} to denote the low-rank prior, the local smoothness in the spatial dimension, and the local smoothness in the spectral/temporal dimension, respectively.

It can be seen that the computational complexities of the three models is with around the similar order of magnitude $\mathcal{O} (n_1n_2^2)$ since $ n_2 $ is generally greater than $ \log(n_1) $, which reflects the relative efficiency of a matrix-based method in dealing with low-rank recovery tasks. Especially, the computational efficiency of our model is comparable to other comparative ones. Considering its more comprehensive encoding of of data priors, it should be rational to say that our method is efficient.
\begin{table}[t]
\fontsize{7}{7}\selectfont
\caption{Comparison of computational complexities and encoded priors of 3DCTV-RPCA, RPCA and LRTV methods.}
\vspace{-4mm}
\centering
\begin{tabular}{c c c c c }
\Xhline{2\arrayrulewidth}
Model & Complexity & L & LSS-S  & LSS-ST\\
\Xhline{1\arrayrulewidth}
3DCTV-RPCA &  $\mathcal{O} (n_1n_2.(\log (n_1)+3n_2))$ & \CheckmarkBold & \CheckmarkBold & \CheckmarkBold \\
\Xhline{1\arrayrulewidth}
LRTV \cite{He2015Total} &  $\mathcal{O} (n_1n_2.(\log (n_1)+n_2))$ & \CheckmarkBold & \CheckmarkBold & \XSolidBrush \\
\Xhline{1\arrayrulewidth}
RPCA \cite{candes2011robust} &  $\mathcal{O} (n_1n_2^2)$ & \CheckmarkBold & \XSolidBrush & \XSolidBrush \\
\Xhline{2\arrayrulewidth}
\end{tabular}
\label{table:complexity}
\vspace{-2mm}
\end{table}

\subsection{Convergence Analysis of Algorithm 1}
Considering that Algorithm 1 is a five-block ADMM, we cannot directly apply the convergence conclusion of the two-block ADMM\cite{eckstein1992douglas, boyd2011distributed} to derive its convergence behavior.  However, owing that each $ \G_i$ is an auxiliary variable of $ \nabla_i \X $, and
each item in the objective function is a convex function, we can give the following convergence result of Algorithm 1.

\begin{theorem}
\label{theorem3}
 \textbf{\emph{The sequence $ (\X^{k},\S^{k},\{ \G_i^k\}_{i=1}^3)$ generated by Algorithm 1 converges to a feasible solution of 3DCTV-RPCA model (\ref{3dnntv_r}), and the corresponding objective function converges to the optimal value $ p^* = \sum_{i=1}^3 \| \G^*\|_* + 3\lambda \| \S^*\|_1 $, where $ (\X^{*},\S^{*},\{ \G_i^*\}_{i=1}^3)$ is an optimal solution of (\ref{3dnntv_r}).}}
\end{theorem}
The proof of Theorem \ref{theorem3} is presented in the supplementary material. Besides this theoretical convergence guarantee, we shall further provide an empirical analysis to support the good convergence behavior of Algorithm 1 in the following subsection.

\subsection{Simulations}
\label{Simulations}
In this section, we will conduct a series of experiments on
synthetic data to test the performance of 3DCTV-RPCA model.
	\begin{figure}[!]
		\centering
		\includegraphics[scale=0.4]{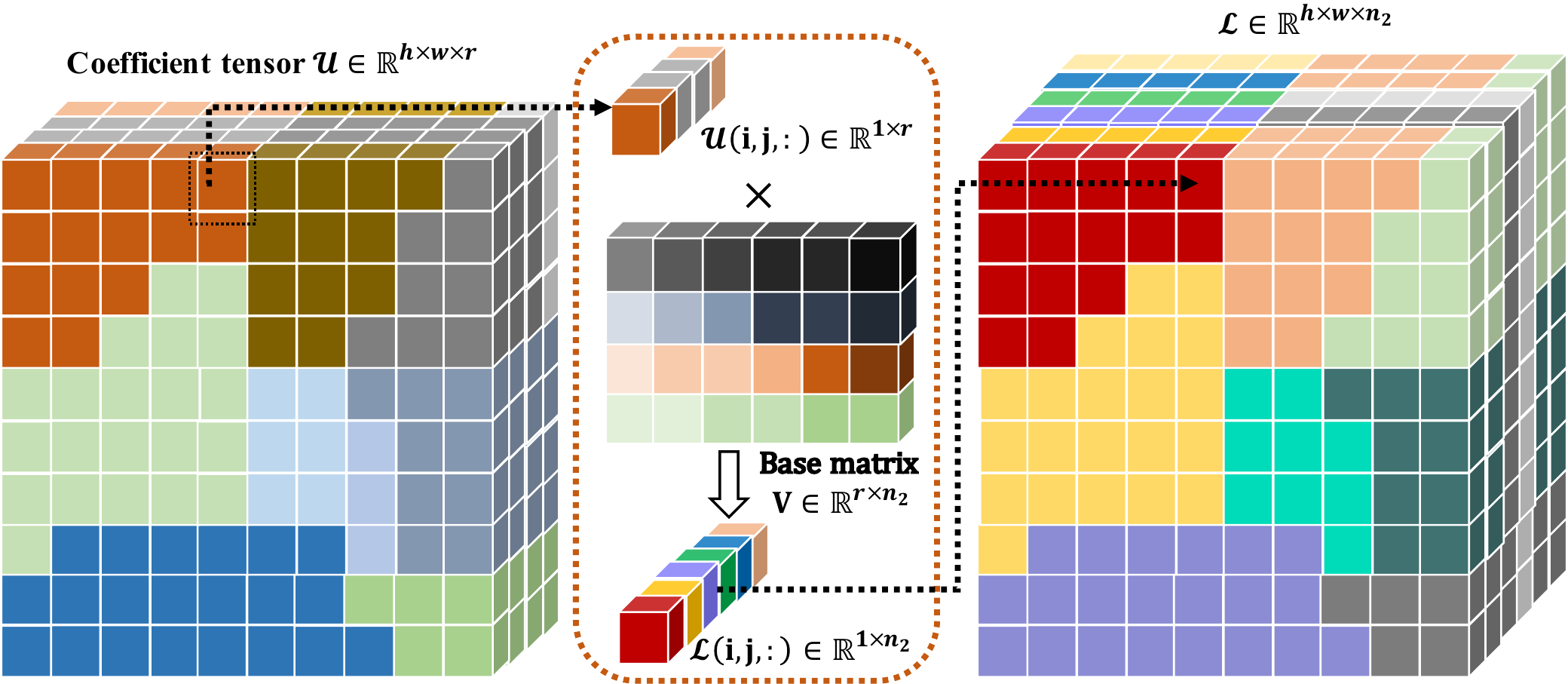}
		\vspace{-3mm}
		\caption{The simulated data generated mechanism of joint low rank and local smoothness data $ \X_0 $.}
		\label{simulation_generate}
		\vspace{-3mm}
	\end{figure}
	
\textbf{Data generation}.  We first generate $ \mathcal{X} =\mathcal{U}\times_3 \V^T $, where the coefficient tensor $\mathcal{U} \in \mathbb{R}^{h\times w \times r} $, the base matrix $ \V \in \mathbb{R}^{r\times n_2}$, and $r \ll \min \{ hw,n_2 \}$. To further make $ \mathcal{X} $ have local-smoothness property, we generate the data in the following manner:
\begin{itemize}
 \item Generating the coefficient tensor $\mathcal{U}$. Precisely, randomly select $r$ initial points, and the remaining points are allocated according to the distance to the initial point. We then divide the space into $r$ regions, each of which has the same representation coefficient vector with entries independently sampled from a $\mathcal{N}(0,1/(hw))$ distribution.
 \item Generating the base matrix $ \V $. Precisely, $\V$ is generated by smoothing each vector $\V(i,:)$ with entries independently sampled from a $\mathcal{N}(0,1)$ distribution.
\end{itemize}
Then, we get $ \X_0=\mbox{unfold}(\mathcal{U}\times_3\V^T) $, the gray value of each column of $ \X_0 $ is normalized into $ \left[ 0, 1\right] $ via the max-min formula, and the entire generation process is shown in Fig. \ref{simulation_generate}. Besides, $ \S_0 $ is generated by choosing a support set $ \Omega $ of size $m$ uniformly at random, and setting $ \S_0 =\mathcal{P}_{\Omega} (\E) $, where $ \E $ is a matrix with independent Bernoulli $ \pm 1 $ entries. Finally, the $\M $ is set as: $\M = \X_0 + \S_0$. In all the following experiments, we set $h=w=20,n_2=200 $.

\textbf{Empirical analysis of convergence}. We provide an empirical analysis for the convergence of Algorithm 1. To this end, we define some variables as the evaluation metrics of algorithm convergence. Precisely, the change at the $k$-th iteration is defined as
\begin{equation}
\mbox{Chg} = \max \{ \mbox{Chg} \M, \mbox{Chg} \X, \mbox{Chg} \S \},
\end{equation}
where Chg$\M=\| \M -\X_{k} -\S_{k} \|_{\infty}$, Chg$\X=\| \X_{k} -\X_{k-1} \|_{\infty}$, Chg$\S=\| \S_{k} -\S_{k-1} \|_{\infty}$.
And the relative errors at the $ k $-th iteration are defined as
\begin{equation}
\begin{split}
\mbox{RelError}\X &=\max \{1, \frac {\| \X_{k-1} -\X_{k} \|_F} {\| \X_{k-1}\|_F} \},\\
\mbox{RelError}\S &=\max \{1, \frac {\| \S_{k-1} -\S_{k} \|_F} {\| \S_{k-1}\|_F} \}, \\
\mbox{RelError}\M &=\max \{1, \frac {\| \M -\X_{k} -\S_{k} \|_F} {\| \M\|_F} \}.
\end{split}
\end{equation}

\begin{figure}[!]
		\centering
		\includegraphics[scale=0.34]{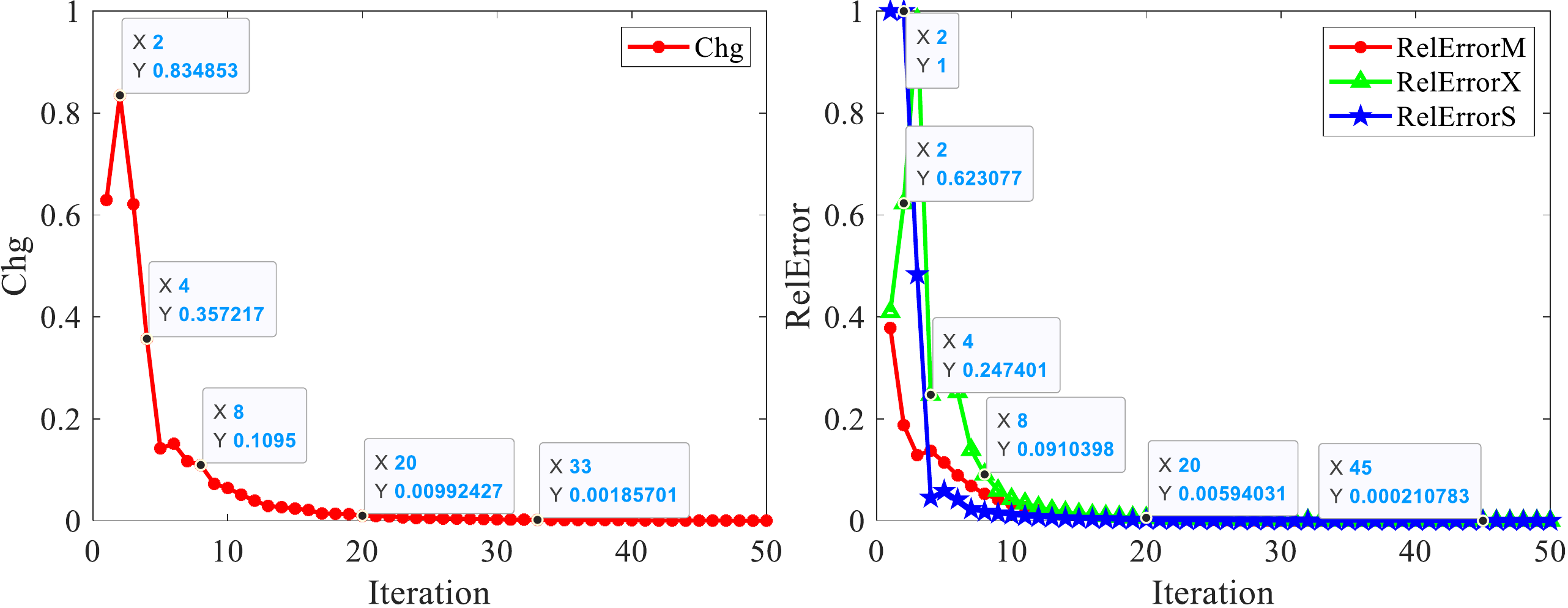}
		\vspace{-6mm}
		\caption{Convergence curves of 3DCTV-RPCA model (\ref{3dnntv_r}).}
		\label{numerical_iteration}
		\vspace{-4mm}
	\end{figure}

In this experiment, we set $ \rho_s = 0.05 $, $ r/n_2=0.1 $. Fig. \ref{numerical_iteration} plots the convergence curves for solving the 3DCTV-RPCA model. We can easily observe that the values of Chg, RelError$\M$, RelError$\S$, and RelError$\X$ decrease rapidly in the first 10 steps, and gradually approach 0 when the number of iterations is greater than 40. This experiment validates the convergence performance of Algorithm 1.

	\begin{figure}[!]
		\centering
		\includegraphics[scale=0.37]{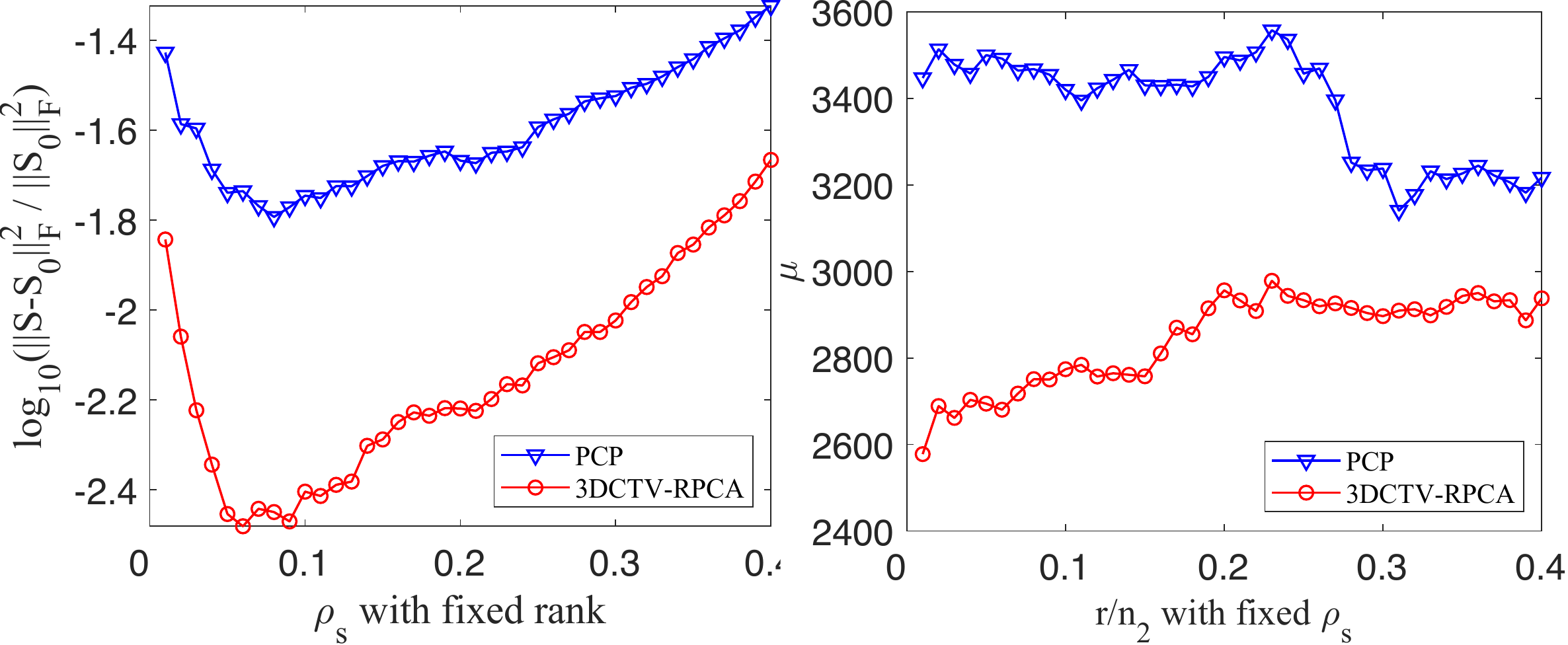}
		\vspace{-7mm}
		\caption{Comparison of logarithmic relative error (left) and the constant value $ \mu $ of incoherence conditions (right) with fixed rank $ r/n=0.05 $ and varying sparsity $ \rho_s $.}
		\label{curve_smoothness}
		\vspace{-4mm}
	\end{figure}
	
	\begin{figure}[!]
		\centering
		\includegraphics[scale=0.37]{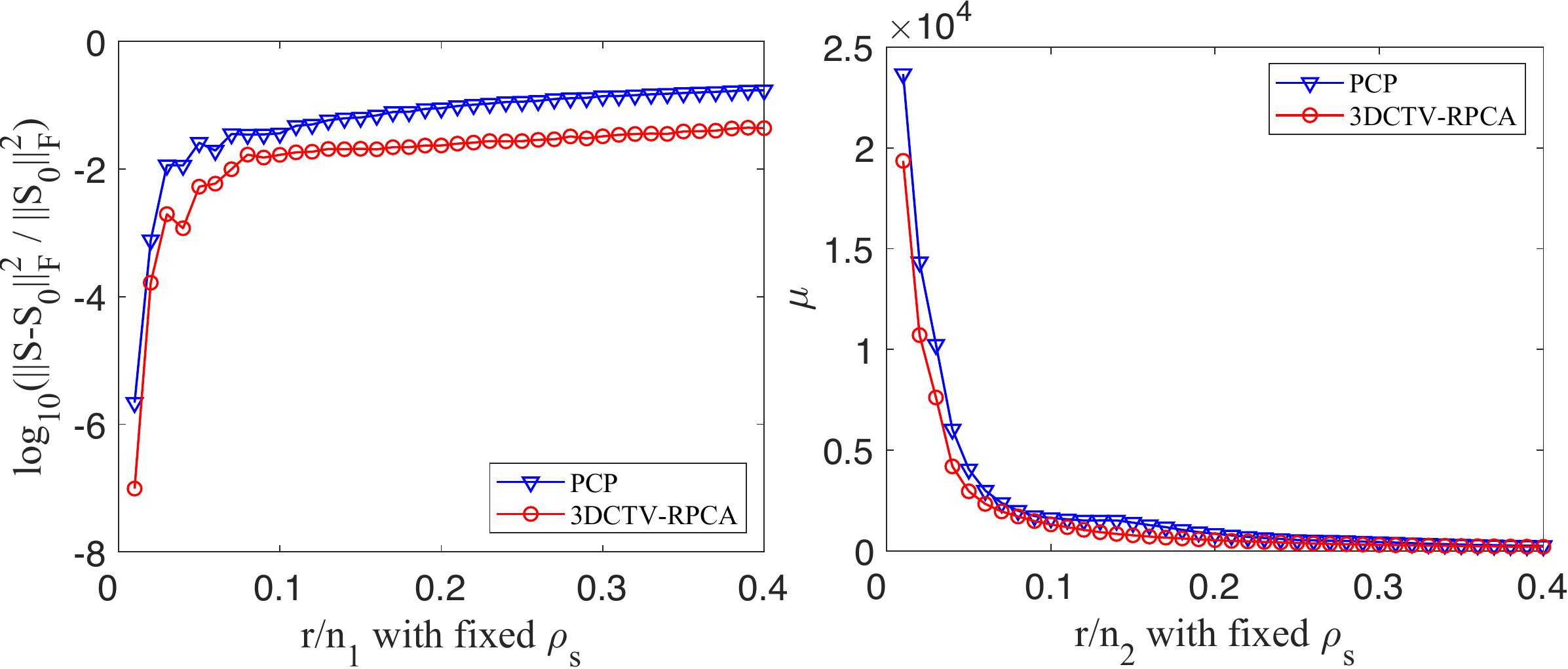}
		\vspace{-7mm}
		\caption{Comparison of logarithmic relative error (left) and the constant value $ \mu $ of incoherence conditions (right) with fixed sparsity $ \rho_s=0.05 $ and varying rank $ r $.}
		\label{curve_smoothness10}
		\vspace{-4mm}
	\end{figure}

\textbf{The RelError$\S $ and the constant value $ \mu $}. To avoid the randomness, we perform 30 times against each test and report the average result. In the following experiments, the logarithmic relative error  $ \log_{10} \mbox{RelError}\S$ of sparse component recovery is used to measure the performance of these algorithms. The logarithmic relative error of $ \S_0 $ is plotted in Figs. \ref{curve_smoothness} and \ref{curve_smoothness10}. From these figures, we can easily see that 3DCTV-RPCA model can always provide smaller logarithmic relative error than PCP model. Besides, we have known that if the constant value $ \mu $ in Eq. (\ref{new_U})-(\ref{new_UV}) is smaller, then the upper bound in Eq. (\ref{pho_r}) is higher. Therefore, a smaller $ \mu $ can be applied to a higher rank value $ \X_0 $, which means that the model can be applied to more matrix recovery tasks. This is because that it is much more difficult to recover high-rank matrices than recover low-rank matrices. Naturally, given the matrix $ \X_0 $, if one model needs a smaller $ \mu $ value, then we can assert that the model needs a weaker incoherence condition. To compare the strength of the incoherence conditions for 3DCTV-RPCA model and PCP model, we denote the respective smallest values of $ \mu $ that need to be met for Eq. (\ref{new_U})-(\ref{new_UV}) for 3DCTV-RPCA model and PCP model by $ \mu (\U_{\scriptsize{\mbox{{g}}}})$, $ \mu (\V_{\scriptsize{\mbox{{g}}}})$, $ \mu (\U_{\scriptsize{\mbox{{g}}}}\V_{\scriptsize{\mbox{{g}}}}^T )$ and $ \mu (\U)$, $ \mu (\V), \mu (\U\V^T )$, respectively. And the smallest constant value $ \mu $ is defined as:
\begin{equation}
\mu (\mbox{3DCTV-RPCA}) = \mbox{max} \{ \mu (\U_{\scriptsize{\mbox{{g}}}}), \mu (\V_{\scriptsize{\mbox{{g}}}}),
\mu (\U_{\scriptsize{\mbox{{g}}}}\V_{\scriptsize{\mbox{{g}}}}^T)  \},
\end{equation}
and
\begin{equation}
\mu (\mbox{PCP}) = \mbox{max} \{ \mu (\U), \mu (\V),\mu (\U\V^T)  \}.
\end{equation}


The curves about constant value $ \mu $ are plotted in Fig. \ref{curve_smoothness} and \ref{curve_smoothness10}. From these figures, we can easily see that the $ \mu $ needed by 3DCTV-RPCA model is smaller than PCP model in almost all cases, which can be seen as the gain brought
by local smoothness. That is, 3DCTV-RPCA model tends to require weaker incoherence condition than PCP model. This is because the CTV regularity encodes the low rank and local smoothness of the original data at the same time, and thus the 3DCTV-RPCA model can provide more accurate prior information in the optimization process of finding the principal components of the simulation data than PCP model.

\begin{table*}[!]
\fontsize{7.5}{7.5}\selectfont
\caption{The quantitative comparison of all competing methods under different levels of noise. The value in the table is the mean value of DC mall dataset, and the best and second results are highlighted in bold italics and underline, respectively.  The value in the table is the average of ten times.}
\vspace{-4mm}
\centering
\begin{tabular}{|c |c |c| c| c| c| c| c| c| c| c| c| c| c|}
\Xhline{2\arrayrulewidth}
Noise &Metric  & 3DCTV-RPCA & PCP & VBRPCA & BRPCA &MPCP & PCPF & MoG & WNNM & RegL1&PRMF & LRTV & LRMR \\
\Xhline{2\arrayrulewidth}
\multicolumn{14}{c}{ Gaussian noise} \cr
\Xhline{2\arrayrulewidth}
\multirow{3}{*}{\makecell[l]{G = 0.10}}
&psnr&\textbf{34.56}&	32.06&	30.74&	29.56 &32.37&	32.61&	\underline{34.22}&	33.02&	32.10&	32.24&	34.10&	33.61 \cr
\cline{2-14}
&ssim& \textbf{0.9629}&	0.9586&	0.9227&	0.8856&	0.9359&	\underline{0.9602}&	0.9505&	0.9569&	0.9228&	0.9342&	0.9543&	0.9570 \cr
\cline{2-14}
&ergas&\textbf{68.71}&	90.69&	106.27&	145.13&	87.46&	84.54&	71.88&	83.38&	92.17&	88.89&	\underline{71.27}&	75.93 \cr
\Xhline{2\arrayrulewidth}
\multirow{3}{*}{\makecell[l]{G = 0.20}}
&psnr&\textbf{30.34}&	27.97&	26.59&	26.23&	27.27&	27.36&	28.36&	29.41&	26.30&	26.92&	\underline{30.20}&	29.49 \cr
\cline{2-14}
&ssim&\textbf{0.9060}&	\underline{0.9001}&	0.8162&	0.8050&	0.8213&	\underline{0.9001}&	0.8409&	0.8921&	0.7806&	0.8174&	0.8931&	0.8996 \cr
\cline{2-14}
&ergas&\textbf{110.62}&	144.80&	169.33&	204.46&	156.38&	137.94&	143.71&	122.99&	181.95&	164.11&	\underline{114.84}&	120.76 \cr
\Xhline{2\arrayrulewidth}
\multirow{3}{*}{\makecell[l]{G = 0.30}}
&psnr& \underline{27.93}&	25.59&	24.28&	23.23&	24.13&	25.85&	24.92&	26.59&	22.84&	23.06&	\textbf{28.07}&	26.90 \cr
\cline{2-14}
&ssim& \underline{0.8457}&	0.8398&	0.7364&	0.7150&	0.7042&	0.8364&	0.7303&	0.8078&	0.6475&	0.6633&	\textbf{0.8460}&	0.8353 \cr
\cline{2-14}
&ergas&\textbf{145.22}&	190.93&	222.97&	254.46&	224.07&	184.46&	214.47&	169.97&	271.98&	256.21&	\underline{150.37}&	162.28 \cr
\Xhline{2\arrayrulewidth}
\multirow{3}{*}{\makecell[l]{G = 0.40}}
&psnr& \underline{26.22}&	23.97&	22.54&	21.83&	21.21&	24.12&	22.45&	24.20&	20.28&	20.79&	\textbf{26.72}&	25.09 \cr
\cline{2-14}
&ssim& 	\underline{0.7852}&	0.7838&	0.6492&	0.6050&	0.5758&	0.7739&	0.6278&	0.7156&	0.5301&	0.5517&	\textbf{0.8005}&	0.7721 \cr
\cline{2-14}
&ergas&\textbf{176.34}&	230.71&	274.29&	304.46&	315.52&	225.61&	288.76&	223.69&	366.99&	335.08&	\underline{188.03}&	199.78 \cr
\Xhline{2\arrayrulewidth}
\multicolumn{14}{c}{ Sparse noise} \cr
\Xhline{2\arrayrulewidth}
\multirow{3}{*}{\makecell[l]{S = 0.10}}
&psnr& \textbf{49.35}&	45.13&	43.16&	38.23&	\underline{47.08}&	45.34&	43.00&	42.28&	43.75&	44.31&	39.29&	41.91\cr
\cline{2-14}
&ssim& \textbf{0.9992}&	0.9988&	0.9979&	0.9663&	\underline{0.9991}&	0.9988&	0.9911&	0.9963&	0.9935&	0.9974&	0.9853&	0.9961 \cr
\cline{2-14}
&ergas&\textbf{15.28}&	27.48&	34.31&	94.58&	\underline{23.26}&	27.14&	93.31&	31.88&	44.07&	24.54&	64.84&	27.37\cr
\Xhline{2\arrayrulewidth}
\multirow{3}{*}{\makecell[l]{S = 0.20}}
&psnr& \textbf{48.26}&	43.49&	42.36&	35.23&	45.17&	\underline{43.76}&	37.59&	41.87&	38.92&	44.19&	39.07&	41.72\cr
\cline{2-14}
&ssim& \textbf{0.9991}&	0.9983&	0.9976&	0.9583&	\underline{0.9987}&	0.9983&	0.9764&	0.9961&	0.9800&	0.9973&	0.9808&	0.9959 \cr
\cline{2-14}
&ergas&\textbf{17.26}&31.58&	36.66&	185.46&	27.50&	30.97&	193.41&	33.59&	144.61&	\underline{24.76}&	107.29&	27.79\cr
\Xhline{2\arrayrulewidth}
\multirow{3}{*}{\makecell[l]{S = 0.30}}
&psnr& \textbf{46.91}&	41.39&	41.41&	33.23&	42.84&	41.72&	37.38&	38.44&	38.83&	\underline{44.03}&	35.64&	41.25\cr
\cline{2-14}
&ssim& \textbf{0.9987}&	0.9973&	0.9973&	0.9463&	\underline{0.9981}&	0.9974&	0.9754&	0.9902&	0.9795&	0.9971&	0.9707&	0.9955 \cr
\cline{2-14}
&ergas& \textbf{19.88}&	38.21&	39.25&	202.64&	32.91&	36.80&	226.66&	50.71&	173.53&	\underline{24.89}&	118.42&	29.49\cr
\Xhline{2\arrayrulewidth}
\multirow{3}{*}{\makecell[l]{S = 0.40}}
&psnr& \textbf{45.12}&	38.63&	40.15&	32.23&	39.95&	39.00&	35.45&	36.05&	35.43&	\underline{43.64}&	35.26&	40.51\cr
\cline{2-14}
&ssim& \textbf{0.9982}&	0.9944&	0.9965&	0.9383&	0.9958&	0.9947&	0.9591&	0.9847&	0.9677&	\underline{0.9967}&	0.9676&	0.9948 \cr
\cline{2-14}
&ergas& \textbf{23.73}&	49.24&	43.00&	228.46&	41.37&	46.45&	391.19&	67.80&	218.80&	\underline{25.33}&	135.16&	31.43\cr
\Xhline{2\arrayrulewidth}
\multicolumn{14}{c}{ Sparse noise and  Gaussian noise} \cr
\Xhline{2\arrayrulewidth}
\multirow{3}{*}{\makecell[l]{G=0.05\\S=0.10}}
&psnr& \textbf{38.34}&	35.31&	34.50&	30.18&	35.58&	35.42&	38.53&	34.92&	\underline{37.09}&	37.04&	36.32&	35.17\cr
\cline{2-14}
&ssim& \textbf{0.9841}&	0.9808&	0.9610&	0.9093&	0.9789&	\underline{0.9810}&	0.9727&	0.9767&	0.9633&	0.9659&	0.9716&	0.9690\cr
\cline{2-14}
&ergas &\textbf{45.02}&	63.29&	70.32&	112.45&	61.95&	62.24&	88.82&	70.37&	\underline{58.95}&	48.85&	63.26&	74.37\cr
\Xhline{2\arrayrulewidth}
\multirow{3}{*}{\makecell[l]{G=0.10\\S=0.10}}
&psnr& \underline{34.02}&	31.40&	29.81&28.39&	32.45&	31.49&	33.14&	32.56&	31.29&	31.36&	\textbf{34.21}&	31.53\cr
\cline{2-14}
&ssim& \textbf{0.9572}&	\underline{0.9533}&		0.8992&0.8635&	0.9512&	0.9529&	0.9232&	0.9518&	0.8849&	0.8977&	0.9480&	0.9341 \cr
\cline{2-14}
&ergas&\textbf{72.87}&	97.78&	86.07&	87.05&	96.51&	98.27&	104.73&	87.28&	95.40&	92.72&	\underline{83.39}&	96.90\cr
\Xhline{2\arrayrulewidth}
\multirow{3}{*}{\makecell[l]{G=0.05\\S=0.20}}
&psnr&\textbf{37.48}&	34.48&	33.99&	26.86&	35.01&	34.60&	35.15&	34.56&	35.84&	35.59&\underline{35.82}&	32.96\cr
\cline{2-14}
&ssim& \textbf{0.9811}&\underline{0.9774}&		0.9544&0.8544&	0.9759&	0.9772&	0.9561&	0.9739&	0.9558&	0.9560&	0.9654&	0.9513 \cr
\cline{2-14}
&ergas&\textbf{49.70}&	69.59&	75.04&	185.33&	66.03&	68.48&	203.53&	72.49&	86.21&	\underline{57.48}&	135.78&	81.87\cr
\Xhline{2\arrayrulewidth}
\multirow{3}{*}{\makecell[l]{G=0.10\\S=0.20}}
&psnr& \textbf{33.22}&	30.65&	29.13&	21.17&	31.72&	30.72&	32.15&	31.96&	31.15&	30.39&	\underline{32.51}&	30.19\cr
\cline{2-14}
&ssim& \textbf{0.9496}&\underline{0.9450}&		0.8821&0.6674&	0.9434&	0.9446&	0.9158&	0.9439&	0.8851&	0.8761&	0.9314&	0.9159 \cr
\cline{2-14}
&ergas&\textbf{79.75}&	106.61&129.46&336.82&	95.02&	105.54&	182.51&	\underline{92.87}&	114.66&	102.83&	120.13&	112.99\cr
\Xhline{2\arrayrulewidth}
\end{tabular}
\label{dcmall evaluation}
\vspace{-4mm}
\end{table*}
	
\begin{figure}[!]
		\centering
		\includegraphics[scale=0.9]{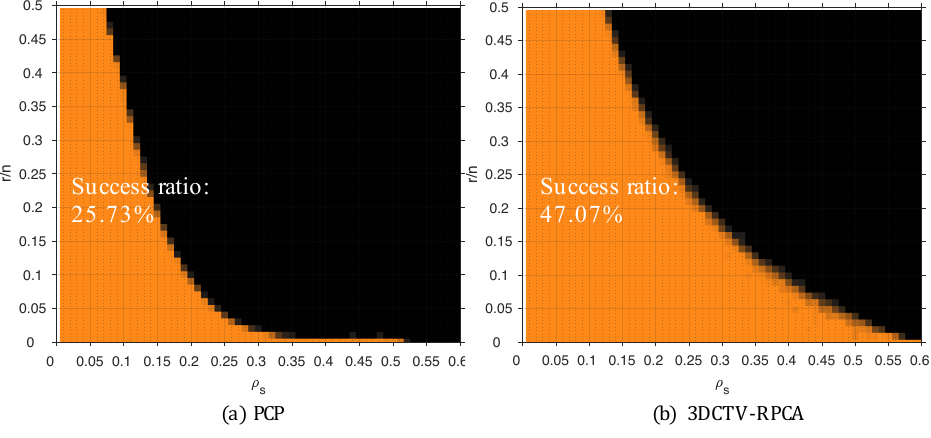}
		\vspace{-4mm}
		\caption{Fraction of correct recoveries across 30 trials, as a function of sparsity of $ \S_0 $ (x-axis) and of rank($ \X_0 $) (y-axis). The phase transition diagram of PCP model (a) and 3DCTV-RPCA model (b). }
		\label{smoothness_pharse}
		\vspace{-6mm}
	\end{figure}

\textbf{Phase transition in sparsity and rank.} We now investigate how the rank of $ \X_0 $ and the sparsity of $ \S_0 $ affect the performance of  3DCTV-RPCA and PCP model. We vary the sparsity $ \rho_s $ of $ \S_0 $ and rank $ r/n $ of $ \X_0 $ between 0 and 0.6, and perform 30 times against each $ (\rho_s,r) $ to avoid the randomness. Here, if the recovered matrix $ \hat{\X} $ satisfies $ \| \X_0 - \hat{\X} \|_F/ \| \X_0 \|_F \leq 0.05$, we assert this trial is a success recovery. The phase transition diagrams are plotted in Fig. \ref{smoothness_pharse}. From the figure, we can easily observe that the 3DCTV-RPCA model can recover more cases, including the cases with higher sparsity and bigger rank, compared to PCP model (\ref{PCP}). Specifically, the percentages of successful recovery area of ​​3DCTV-RPCA and PCP are 47.07\% and 25.71\%, respectively. It is can substantiated that the 3DCTV-RPCA model (\ref{3dnntv_r}) can better deal with the joint low rank and local smoothness data than PCP model (\ref{PCP}).

\section{Extensional Applications}
We further test the performance of the 3DCTV-RPCA model (\ref{3dnntv_r}) on three applications, including hyperspectral image (HSI) denoising, multispectral image (MSI) denoising, and background modeling from surveillance video.

\subsection{Hyperspectral Image Denoising}
Compared to traditional image systems, a HSI consists of various intensities representing the radiance's integrals captured by sensors over various discrete bands. A HSI data has a strong correlation across its spectral mode, that is, low rankness property. Through utilizing this property, a series of studies showed good performance in various HSI related tasks \cite{plaza2009recent}, e.g., classification \cite{Zhao2016High}, super-resolution \cite{xie2020mhf}, compressive sensing \cite{peng2020enhanced} and unmixing \cite{yao2019nonconvex}.

We choose some classic and state-of-the-art \textbf{L+S} matrix decomposition methods as the competing methods, including RPCA\cite{candes2011robust} \footnote{\url{https://github.com/dlaptev/RobustPCA}} that used to solve the PCP model (\ref{PCP}), BRPCA \cite{ding2011bayesian} \footnote{\url{http://people.ee.duke.edu/lcarin/BCS.html}}, VBRPCA\cite{babacan2012sparse} \footnote{\url{http://www.dbabacan.info/publications.html}}, MPCP \cite{zhan2015robust}, PCPF \cite{chiang2016robust}, RegL1 \cite{zheng2012practical} \footnote{\url{https://sites.google.com/site/yinqiangzheng/}}, WNNM \cite{Gu2014Weighted} \footnote{\url{http://gr.xjtu.edu.cn/web/dymeng/2}}, PRMF\cite{wang2012probabilistic} \footnote{\url{http://winsty.net/prmf.html}}, MOG\cite{meng2013robust} \footnote{\url{http://gr.xjtu.edu.cn/web/dymeng/2}}, OMoGMF\cite{yong2017robust} and LRMR \cite{Zhang2014Hyperspectral}. Since LRTV\cite{He2015Total, shi2015lrtv} achieves state-of-the-art performance among all methods based on the \textbf{L\& LSS+S} form, we choose LRTV as the representative method.

The selected data is pure DC mall dataset\cite{Zhang2014Hyperspectral}, whose size is $200\times 200\times 160$ with complex structure and texture. We reshape each band as a vector and stack all the vectors to form a matrix, resulting in the final data matrix with size $40000\times 160 $. Before conducting this experiment, the gray value of each band was normalized into $[0, 1]$ via the max-min formula.

In TABLE \ref{dcmall evaluation}, we list the denoising results on three different types of noise, including Gaussian noise with zero-mean and different standard deviation, sparse noise with different percentages, and mixed noise with Gaussian noise and sparse noise. Specifically, "G" and "S" represent Gaussian and sparse noise, respectively, and their value is the degree of noise. From the table, we can easily find that the 3DCTV-RPCA model achieves the best performance among almost all the competing methods in all noise cases except the second best in two Gaussian noise cases.

\begin{figure*}[!]
		\centering
		\includegraphics[scale=0.8]{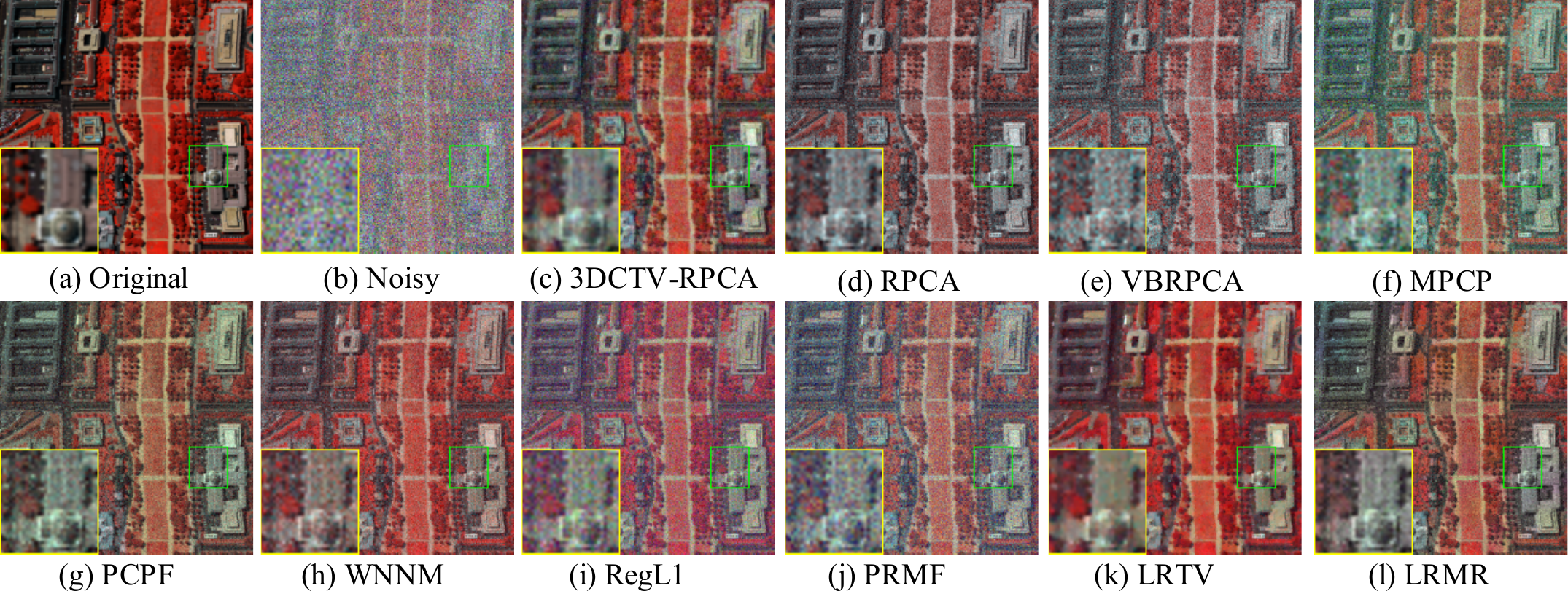}
		\vspace{-5mm}
		\caption{Recovered images of all competing methods with bands 58-27-17 as R-G-B. (a) The simulated DC mall image. (b) The noise image with Gaussian noise variance is 0.4. (c-i) The results were obtained by all comparison methods, with a demarcated zoomed in three times for easy observation.}
		\vspace{-3mm}
		\label{dcmall_G04}
	\end{figure*}

\begin{figure*}[!]
		\centering
		\includegraphics[scale=0.8]{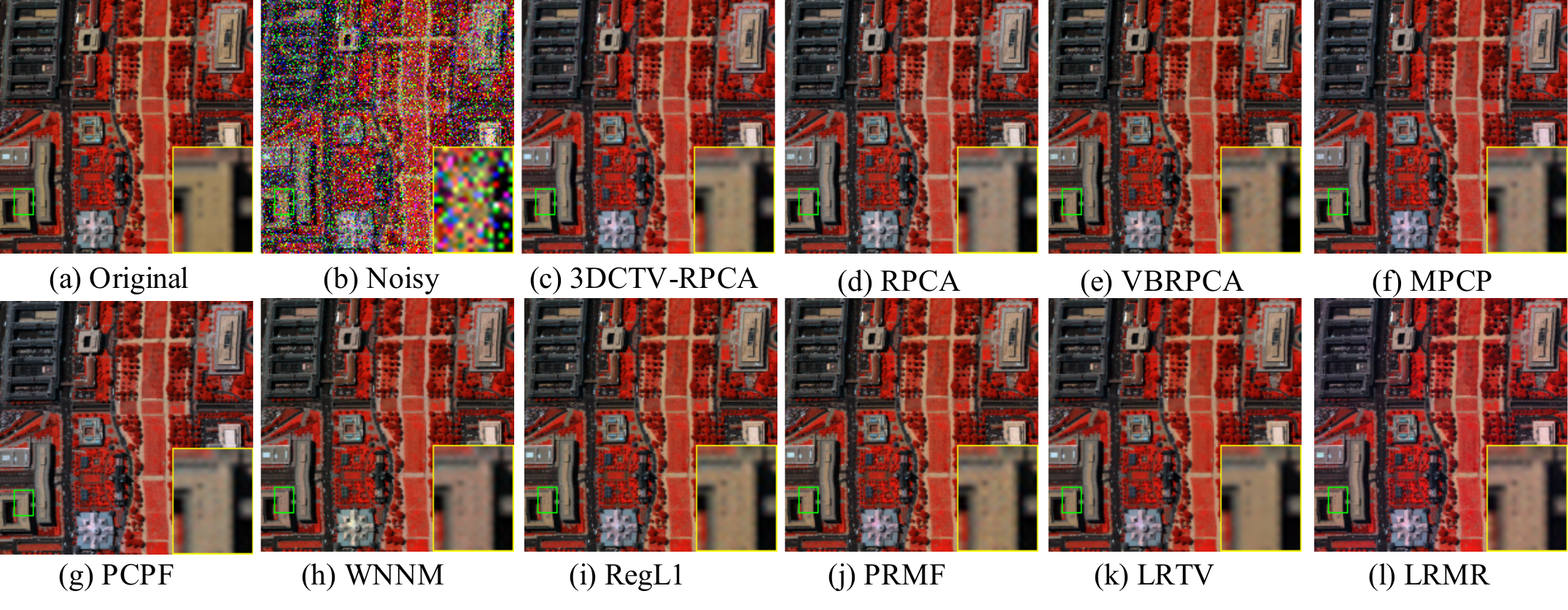}
		\vspace{-5mm}
		\caption{Recovered images of all competing methods with bands 58-27-17 as R-G-B. (a) The simulated DC mall image. (b) The noise image with Gaussian noise variance is 0.05 and the sparse noise variance is 0.2. (c-i) Restoration results obtained by all comparison methods, with a demarcated zoomed in 3 times for easy observation. }
		\vspace{-3mm}
		\label{dcmall_show}
	\end{figure*}

To better visualize comparison, we choose three bands of HSI to form a pseudo-color image to show all competing methods' visual restoration performance. The image restorations of all methods under Gaussian noise and mixture noise are plotted in Fig. \ref{dcmall_G04} and Fig. \ref{dcmall_show}, respectively. It's easy to see that the 3DCTV-RPCA model can achieve better noise removal performance in all cases, i.e., more faithfully maintaining the image's color fidelity and smoothness property. It should be noted that if we choose the $ \ell_2 $-norm to characterize the distribution of pure Gaussian noise, the CTV regularization can also obtain a good performance. Thus, we also provide the comparison results for combining the different noise terms with CTV regularization in the supplementary materials.

\begin{figure*}[!]
		\centering
		\includegraphics[scale=0.8]{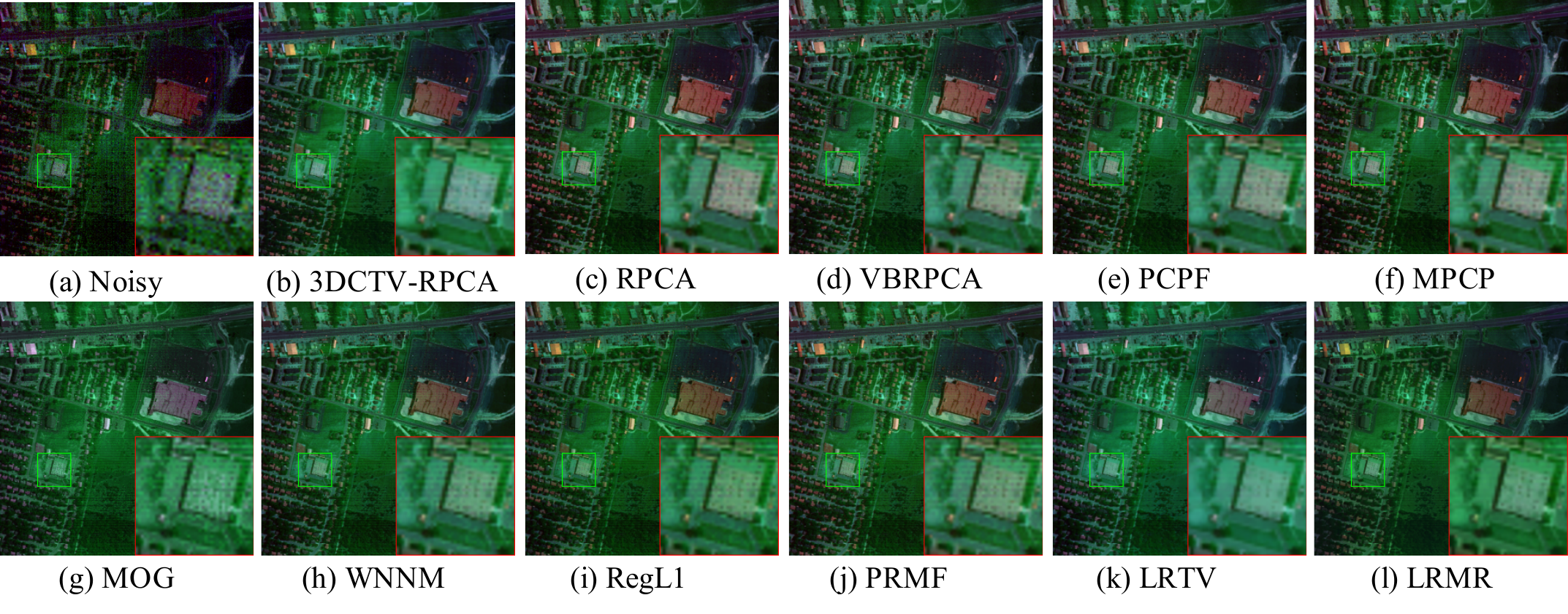}
		\vspace{-3mm}
		\caption{Recovered images of all competing methods with bands 6-104-36 as R-G-B. (a) The original urban part image. (b-l) Restoration results obtained by 11 comparison methods, with a demarcated zoomed in three times for easy observation. }
		\label{urban_show}
	\end{figure*}

Furthermore, we test all competing method's performances on real urban part data.  We select the sub-figure of urban part data with the size of $200\times 200\times 210$, and reshape each band as a vector and stack all the vectors to form a matrix, resulting in the final data matrix with size $40000\times 210 $. The pseudo-color image is provided in Fig. \ref{urban_show}. In the figure, it is easy to see that all competing methods have not finely removed the complicated noise cleanly except for our proposed 3DCTV-RPCA model and LRTV model. The 3DCTV-RPCA model better balances the low-rankness and local smoothness of HSI data to get a good denoising effect with better maintaining the local texture information of the original image. While when LRTV removes the noise, it inclines to bring excessive smoothness phenomena and result in certainly hampering the color fidelity and local texture of an image.
	
\begin{figure*}[!]
		\centering
		\includegraphics[scale=0.58]{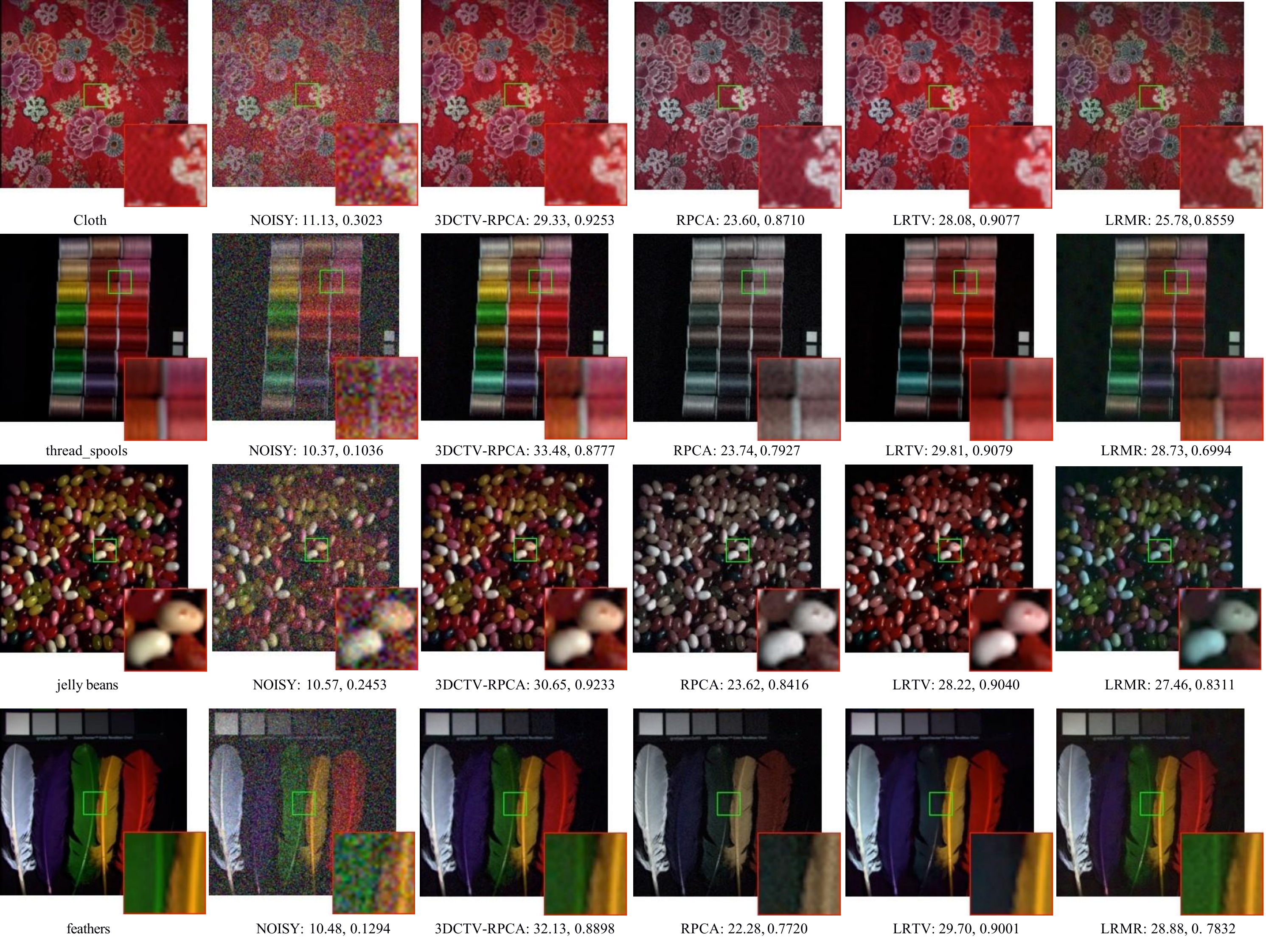}
		\vspace{-4mm}
		\caption{From upper to lower: typical MSI scenes from cloth, thread spools,  jelly beans, and feathers datasets, pseudo-color image (R: 23, G: 13, B: 4) of the original image, and the restored images recovered by RPCA, 3DCTV-RPCA, LRTV, and LRMR. }
		\label{msi_show}
		\vspace{-4mm}
	\end{figure*}
	\begin{figure}[!]
		\centering
		\includegraphics[scale=0.58]{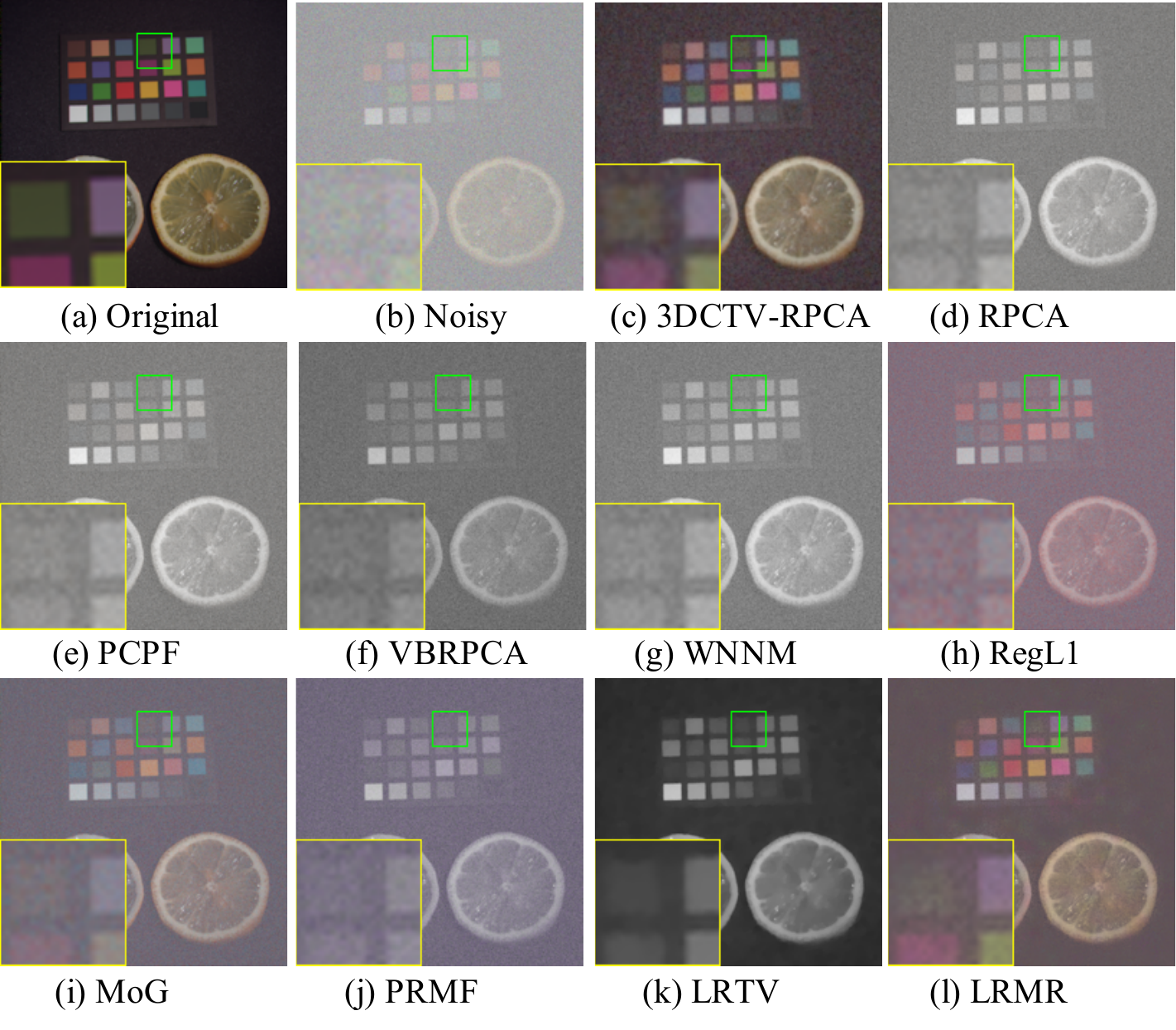}
		\vspace{-4mm}
		\caption{ Recovered images of all competing methods with bands 23-13-4 as R-G-B. (a) The simulated lemon-slices-ms image in CAVE datasets. (b) The noise image with Gaussian noise variance is 0.2. (c-l) Restoration results obtained by ten comparison methods with a demarcated zoomed in 3.5 times for easy observation.}
		\label{CAVE_G2}
	\end{figure}
	
	\begin{figure}[!]
		\centering
		\includegraphics[scale=0.58]{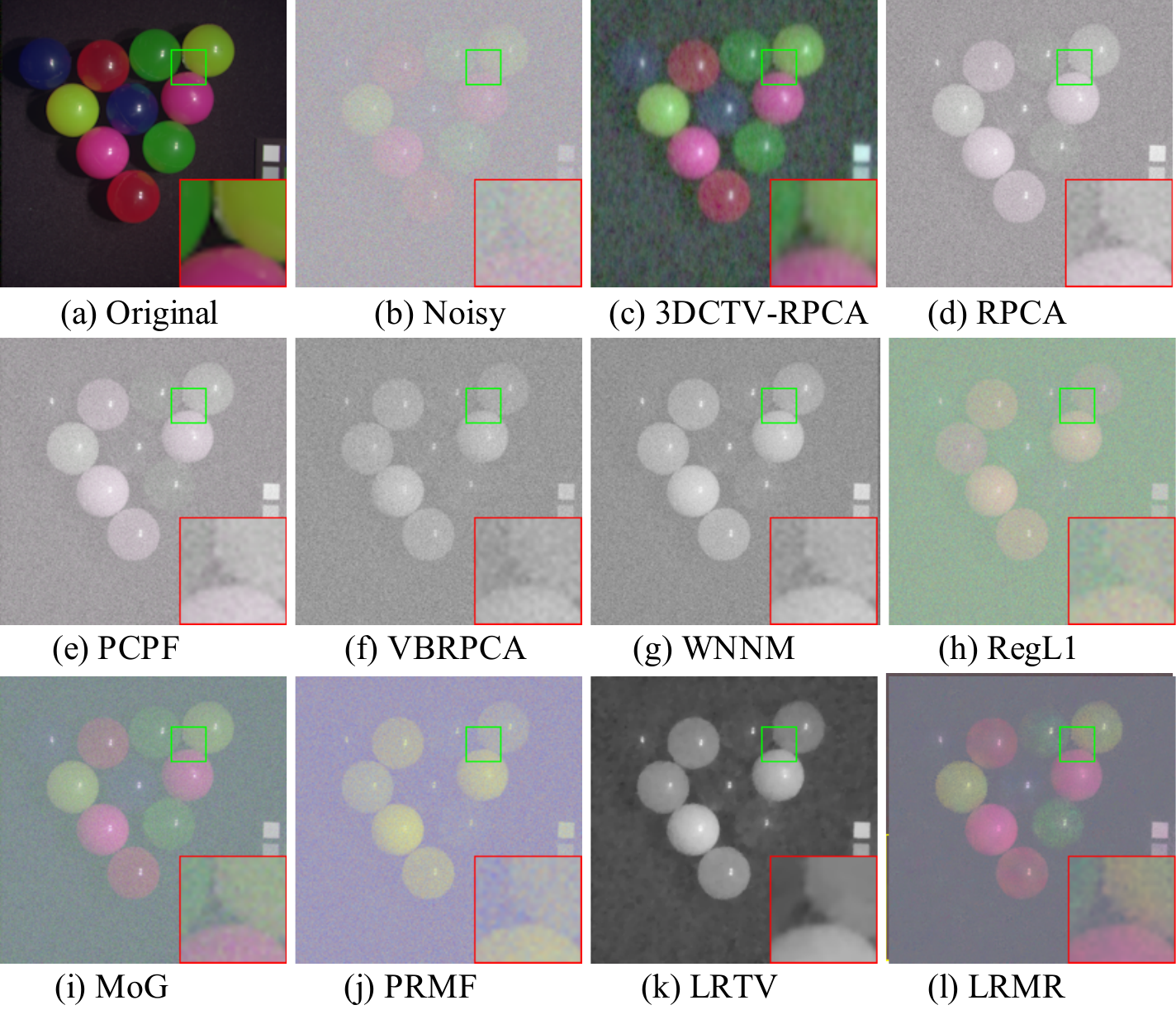}
		\vspace{-4mm}
		\caption{Recovered images of all competing methods with bands 23-13-4 as R-G-B. (a) The simulated superballs-ms image in CAVE datasets. (b) The noise image with Gaussian noise variance is 0.4. (c-l) Restoration results obtained by ten comparison methods with a demarcated zoomed in 3.5 times for easy observation.}
		\label{CAVE_G4}
	\end{figure}

\subsection{Multispectral Image Denoising}
MSI is similar to HSI, but its imaging spectrum in the visible range is oftentimes relatively smaller than HSI. In this experiment, we use the CAVE database \footnote{\url{https://www.cs.columbia.edu/CAVE/databases/multispectral/}}, which was first proposed in  \cite{yasuma2010generalized} and then was generally considered as benchmark database for MSI data processing tasks. We use the same noise settings as the previous HSI denoising task. And all competing methods' parameter settings are followed the suggested settings of their original literatures.

In TABLE \ref{MSI evaluation}, we list the average results over 10 independent trials in terms of three evaluation indices. It is easy to see that 3DCTV-RPCA model get better performance among all the competing methods. Specifically, for Gaussian noise, the recovery effect of the  3DCTV-RPCA model is slightly higher than that of LRTV, which ranks the second among all the competing methods. For sparse noise and mixed noise, 3DCTV-RPCA ranks the first with a more evident performance gain. Though the 3DCTV-RPCA model only incorporates the local smoothness property into the PCP model, it can significantly improve the repair effect of the PCP model (\ref{PCP}).

In Fig. \ref{msi_show},  we select some well-performed techniques and provide the restored images obtained by them for visual comparison. This figure clearly shows that 3DCTV-RPCA model can maintain better texture details of the image and achieve better color fidelity, which is consistent with the evaluation indices listed in TABLE \ref{MSI evaluation}.

We further demonstrate some restorations of competing methods for visual comparison under large noise variance in Fig. \ref{CAVE_G2}, \ref{CAVE_G4}. From the figures, we can see that all methods except the 3DCTV-RPCA model, have obviously damaged the color fidelity of pseudo-color pictures. Comparatively, 3DCTV-RPCA model still maintains a relatively better performance. Note that if the noise is pure Gaussian distribution, we can replace the $ \ell_1$-norm with the $ \ell_2$-norm in 3DCTV-RPCA model (\ref{3dnntv_r}) to further improve the performance. More details about such extension model can be found in the supplementary materials.

\begin{table*}[htbp]
\fontsize{7.5}{7.5}\selectfont
\caption{The quantitative comparison of all competing methods under different noise levels on 32 scenes in the CAVE database. Each value is the mean of all data performance. The best and second results on each line are highlighted in bold italics and underline, respectively. The value in the table is the average of ten times.}
\vspace{-4mm}
\centering
\begin{tabular}{|c |c |c| c| c| c| c| c| c| c| c| c| c|}
\Xhline{2\arrayrulewidth}
Noise &Metric  & Noisy & 3DCTV-RPCA & PCP & VBRPCA & BRPCA  & RegL1 & WNNM & PRMF &MoG & LRTV & LRMR \\
\Xhline{2\arrayrulewidth}
\multicolumn{13}{c}{ Gaussian noise} \cr
\Xhline{2\arrayrulewidth}
\multirow{3}{*}{\makecell[l]{G=0.10}}
&psnr&20.00&	\underline{33.47}&	25.59&	25.97&	24.95&	27.75&	26.87&	27.26&	29.24&	\textbf{33.61}&	32.88\cr
\cline{2-13}
&ssim&0.4180&\underline{0.9006}&	0.8450&	0.7416&	0.7060&	0.7703&	0.7960&	0.7654&	0.8176&	\textbf{0.9165}&	0.8576\cr
\cline{2-13}
&ergas&520.58&\textbf{111.16}&	282.86&	263.68&	325.45&	214.12&	250.88&	227.62&	181.48&	120.64&	\underline{119.90}\cr
\Xhline{2\arrayrulewidth}
\multirow{3}{*}{\makecell[l]{G=0.20}}
&psnr&13.98&	\underline{30.27}&	22.44&	22.04&	20.92&	22.28&	23.45&	22.01&	24.29&	\textbf{30.60}&	28.55\cr
\cline{2-13}
&ssim&0.2033&\underline{0.7803}&	0.7185&	0.5529&	0.4048&	0.5194&	0.6325&	0.5266&	0.6175&	\textbf{0.8601}&	0.7087\cr
\cline{2-13}
&ergas&1041.18&\textbf{158.94}&	385.78&	400.42&	534.78&	405.52&	350.89&	409.77&	319.12&	\underline{182.86}&	191.38\cr
\Xhline{2\arrayrulewidth}
\multirow{3}{*}{\makecell[l]{G=0.30}}
&psnr&10.46&	\textbf{28.24}&	20.54&	19.70&	15.97&	18.85&	21.16&	18.73&	21.08&	\underline{27.94}&	25.80\cr
\cline{2-13}
&ssim&0.1246&	\underline{0.6712}&	0.6248&	0.4313&	0.3570&	0.3686&	0.5199&	0.3620&	0.4665&	\textbf{0.8067}&	0.6023\cr
\cline{2-13}
&ergas&1561.74&\textbf{199.46}&	471.91&	518.96&	698.66&	609.28&	444.93&	595.52&	464.84&	268.52&	\underline{257.73}\cr
\Xhline{2\arrayrulewidth}
\multirow{3}{*}{\makecell[l]{G=0.40}}
&psnr&7.96&\textbf{26.69}&	19.25&	18.03&	14.87&	16.43&	19.31&	16.24&	18.76&	\underline{26.29}&	23.59\cr
\cline{2-13}
&ssim&0.0847&\underline{0.5778}&	0.5533&	0.3443&	0.1982&	0.2758&	0.4394&	0.2463&	0.3670&	\textbf{0.7651}&	0.5139\cr
\cline{2-13}
&ergas&2082.38&\textbf{237.36}&	543.23&	625.94&	906.17&	807.23&	559.56&	799.78&	609.39&	349.82&	\underline{329.15}\cr
\Xhline{2\arrayrulewidth}
\multicolumn{13}{c}{ Sparse Noise} \cr
\Xhline{2\arrayrulewidth}
\multirow{3}{*}{\makecell[l]{S=0.10}}
&psnr&13.79&	\textbf{41.62}&	30.17&	35.45&	34.34&	35.14&	33.70&	34.97&	32.76&	36.28&	\underline{38.57}\cr
\cline{2-13}
&ssim&0.2097&	\textbf{0.9967}&	0.9701&	0.9834&	0.9332&	0.9471&	0.9619&	0.9647&	0.9336&	0.9683&	\underline{0.9722}\cr
\cline{2-13}
&ergas&1085.06&\textbf{	54.75}&	217.07&	111.78&	173.16&	156.43&	154.40&	123.26&	208.10&	104.77&	\underline{86.08}\cr
\Xhline{2\arrayrulewidth}
\multirow{3}{*}{\makecell[l]{S=0.20}}
&psnr&10.78&\textbf{	40.57}&	27.79&	33.92&	31.10&	32.47&	32.18&	34.18&	31.27&	35.76&	\underline{36.36}\cr
\cline{2-13}
&ssim&0.1277&	\textbf{0.9958}&	0.9513&	\underline{0.9745}&	0.9115&	0.9215&	0.9509&	0.9554&	0.9142&	0.9642&	0.9598\cr
\cline{2-13}
&ergas&1535.01&	\textbf{60.17}&	263.17&	127.63&	189.85&	236.45&	166.64&	132.73&	339.41&	113.18&	\underline{93.20}\cr
\Xhline{2\arrayrulewidth}
\multirow{3}{*}{\makecell[l]{S=0.30}}
&psnr&9.02&	\textbf{39.42}&25.32&	31.99&	24.31&	28.98&	30.35&	32.22&	27.83&	\underline{34.45}&	33.63\cr
\cline{2-13}
&ssim&0.0903&	\textbf{0.9945}&	0.9223&	0.9591&	0.8726&	0.8720&	0.9247&	0.9293&	0.8254&	\underline{0.9529}&	0.9127\cr
\cline{2-13}
&ergas&1879.68&	\textbf{66.82}&	323.36&	151.94&	375.31&	371.08&	188.83&	170.41&	421.21&	149.37&	\underline{138.36}\cr
\Xhline{2\arrayrulewidth}
\multirow{3}{*}{\makecell[l]{S=0.40}}
&psnr&7.77&	\textbf{36.00}&	22.81&	29.57&	22.73&	25.85&	28.23&	28.83&	20.42&	\underline{32.44}&	30.94\cr
\cline{2-13}
&ssim&0.0669&	\textbf{0.9883}&	0.8789&	0.9290&	0.8400&	0.7828&	0.8805&	0.8672&	0.5759&	\underline{0.9258}&	0.8643\cr
\cline{2-13}
&ergas&2170.82&\textbf{	92.35}&	401.81&	190.36&	515.82&	467.02&	225.16&	219.89&	715.15&	211.86&	\underline{170.61}\cr
\Xhline{2\arrayrulewidth}
\multicolumn{13}{c}{ Sparse Noise and Gaussian noise} \cr
\Xhline{2\arrayrulewidth}
\multirow{3}{*}{\makecell[l]{G = 0.05 \\ S = 0.10}}
&psnr&13.61&	\textbf{36.19}&	27.09&	29.83&	25.13&	31.20&	29.98&	31.24&	31.16&	\underline{34.84}&	33.69\cr
\cline{2-13}
&ssim&0.1977&\textbf{	0.9547}&	0.8999&	0.8804&	0.6089&	0.8751&	0.8818&	0.8850&	0.8285&	\underline{0.9447}&	0.8466\cr
\cline{2-13}
&ergas&1105.10&	\textbf{83.87}&	256.85&	179.19&	282.10&	178.67&	186.15&	151.61&	196.73&	108.73&	\underline{109.39}\cr
\Xhline{2\arrayrulewidth}
\multirow{3}{*}{\makecell[l]{G = 0.05 \\ S = 0.20}}
&psnr&10.70&	\textbf{35.57}&	25.61&	29.17&	18.51&	29.33&	29.31&	30.10&	30.44&	\underline{34.31}&	32.05\cr
\cline{2-13}
&ssim&0.1240&	\textbf{0.9488}&	0.8765&	0.8566&	0.3373&	0.8403&	0.8624&	0.8534&	0.8334&	\underline{0.9368}&	0.8116\cr
\cline{2-13}
&ergas&1547.20&	\textbf{90.10}&	299.23&	192.06&	618.28&	280.38&	195.34&	170.68&	285.87&	\underline{116.51}&	133.00\cr
\Xhline{2\arrayrulewidth}
\multirow{3}{*}{\makecell[l]{G = 0.10 \\ S = 0.10}}
&psnr&13.18 &	\underline{33.50}&	24.76&	26.34&	23.50&	27.81&	27.16&	27.36&	27.94&	\textbf{33.67}&	29.77\cr
\cline{2-13}
&ssim&0.1764 &	\underline{0.8967}&	0.8220&	0.7753&	0.5283&	0.7515&	0.7806&	0.7494&	0.7097&	\textbf{0.9108}&	0.7456\cr
\cline{2-13}
&ergas&1155.60&	\textbf{111.29}&	309.33&	253.87&	340.62&	222.96&	237.39&	223.46&	229.08&	\underline{126.79}&	170.02\cr
\Xhline{2\arrayrulewidth}
\multirow{3}{*}{\makecell[l]{G = 0.10 \\ S = 0.20}}
&psnr&10.50&	\underline{32.99}&	23.62&	25.75&	17.87&	26.38&	26.48&	26.13&	27.17&	\textbf{33.04}&	28.75\cr
\cline{2-13}
&ssim&0.1156&	\underline{0.8871}&	0.7972&	0.7488&	0.3209&	0.7107&	0.7515&	0.6988&	0.7077&	\underline{0.9034}&	0.7173\cr
\cline{2-13}
&ergas&1580.10&\textbf{	118.07}&	350.30&	270.81&	669.86&	304.69&	251.84&	254.89&	316.60&	\underline{135.76}&	190.70\cr
\Xhline{2\arrayrulewidth}
\end{tabular}
\label{MSI evaluation}
\end{table*}
	
\begin{table*}[htbp]
\fontsize{9}{9}\selectfont
\caption{AUC comparison of all competing methods on all video sequences in the Li dataset. Each value is averaged over all foreground-annotated frames in the corresponding video. The most right column lists the average performance of each competing method overall video sequences. The best and second results in each video sequence are highlighted in bold italics and underline, respectively. The value in the table is the average of ten times.}
\vspace{-4mm}
\centering
\begin{tabular}{c c c c c c c c c c c}
\Xhline{3\arrayrulewidth}
\multirow{2}{*}{\makecell[l]{Methods}}
& \multicolumn{10}{c}{data} \\
\cline{2-11}
&  airp.&	boot.	&shop.	&lobb.	&esca.	&curt.	&camp.&	wate.&	foun.	&Average \\
\Xhline{2\arrayrulewidth}
PCP\cite{candes2011robust} &0.8721&	0.9168&	0.9445&	0.9130&	0.9050&	0.8722&	0.8917&	0.8345&	\underline{0.9418}&	0.8991 \\
3DCTV-RPCA &0.9178&	0.9107&\textbf{0.9541}&	\textbf{\emph{0.9337}}&	\textbf{0.9160}&	0.8710&	0.8814&	0.9386&	0.9383&	\underline{0.9180} \\
MPCP \cite{zhan2015robust} & \underline{0.9363}&	\textbf{\emph{0.9298}}&	0.9472&	\underline{0.9315}&	\underline{0.9144}&	\underline{0.9510}&	\underline{0.8919}&	\underline{0.9651}&	\textbf{\emph{0.9440}}&	\textbf{0.9347} \\
PCPF \cite{chiang2016robust} & \textbf{0.9367}&	0.9242&	0.8987&	0.8352&	0.9022&	\textbf{0.9572}&	0.8458&	\textbf{0.9729}&	0.8593&	0.9036 \\
GODEC \cite{zhou2011godec} & 0.9001&	0.9046&	0.9187&	0.8556&	0.9125&	0.9131&	0.8693&	0.9370&	0.9099&	0.9023\\
DECOLOR \cite{zhou2012moving} & 0.8627&	0.8910&	0.9462&	0.9241&	0.9077&	0.8864&	\textbf{0.8945}&	0.8000&	0.9443&	0.8952 \\
OMoGMF \cite{yong2017robust} & 0.9143&	0.9238&	\underline{0.9478}&	0.9252&	0.9112&	0.9049&	0.8877&	0.8958&	0.9419&	0.9170 \\
RegL1 \cite{zheng2012practical}  & 0.8977&	\underline{0.9249}&	0.9423&	0.8819&	0.4159&	0.8899&	0.8871&	0.8920&	0.9194&	0.8501 \\
PRMF\cite{wang2012probabilistic} & 0.8905&	0.9218&	0.9415&	0.8818&	0.9065&	0.8806&	0.8865&	0.8799&	0.9166&	0.9006 \\
\Xhline{3\arrayrulewidth}
\end{tabular}
\label{table:Back_Fore}
\end{table*}

\subsection{Background Modeling from Surveillance Video}
This task is a traditional online robust PCA task, aiming at decomposing a sequence of surveillance video into its foreground and background components. The former is generally modeled with \textbf{S} prior, and the latter is \textbf{L+LSS} prior. Since we can get some extra data before the next batch of data arrives, some methods based on subspace information embedding, such as MPCP and PCPF models, will get better performance on this task. To apply PCPF and MPCP models, here we choose 20 clean frames that contain no or tiny foreground objects as additional data to extract the subspace knowledge.

In this experiment, we employ the Li dataset \footnote{\url{http://perception.i2r.a-star.edu.sg/bk model/bk index.html}} as the benchmark. This dataset contains nine video sequences, and each frame in video sequence was obtained under a fixed camera to some certain scene. These video sequences range over a wide range of background cases, like static background (e.g., airport, bootstrap, shopping mall), illumination changes (e.g., lobby), dynamic background indoors (e.g., escalator, curtain), and dynamic background outdoors (e.g., campus, water-surface, fountain).

\begin{figure*}[!]
		\centering
		\includegraphics[scale=0.44]{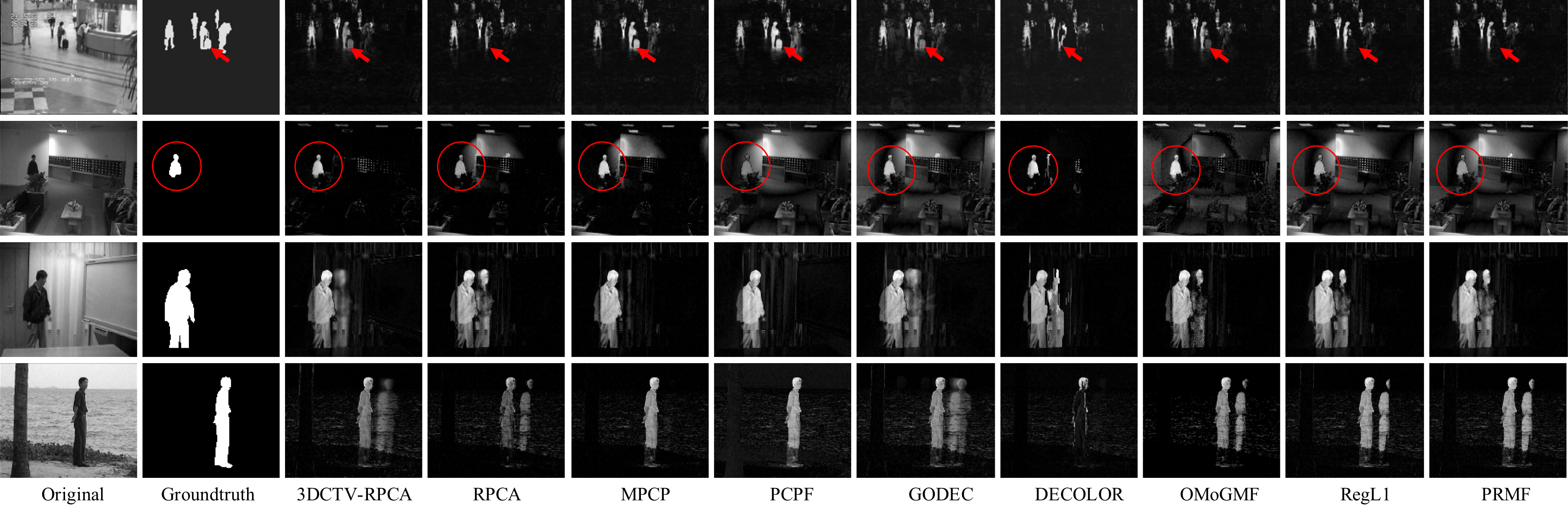}
		\vspace{-4mm}
		\caption{Visual restoration effect display of all comparison methods. From left to right: the original frames, the ground-truths of foreground objects, those detected by all competing methods.}
		\label{back_fore_sep}
		\vspace{-6mm}
	\end{figure*}

In TABLE \ref{table:Back_Fore}, we can see that 3DCTV-RPCA, and MPCP model get relatively better performance than others. Although 3DCTV-RPCA model is 0.016 lower in AUC than MPCP, considering that 3DCTV-RPCA model does not require additional data to provide subspace information and can be readily applied, it thus should be rational to say that our method is still with certain superiority. As compared to PCP model, 3DCTV-RPCA model gains 0.019 better performance in AUC, which also confirms the effect of introducing local smoothness priors in the PCP model (\ref{PCP}), thus demonstrating the effectiveness of the 3DCTV-RPCA model. Furthermore, We show the visual separation performance of all competing methods in Fig.\ref{back_fore_sep}. From the figure, it is easy to see that the 3DCTV-RPCA model can be less affected by other components and thus tends to get good performance. In a word, we can assert that both 3DCTV-RPCA and MPCP model obtain relatively better performance than other competing methods.

\section{Conclusion}
In this work, we have considered the following problem: can we make a matrix decomposition in terms of \textbf{L \& LSS + S} form exactly? To address this issue, we have proposed an ameliorated RPCA model named 3DCTV-RPCA by fully exploiting and encoding the prior information underlying such joint low-rank and local smoothness matrices. Specifically, using a modification of Golfing scheme, we prove that under some suitable assumptions, our model can decompose both components exactly, which should be the first theoretical guarantee among all such related methods combining low rankness and local smoothness. A series of experiments on simulations and real applications are carried out to demonstrate
the general validity of the proposed 3DCTV-RPCA model. In addition, 
more extended application of 3DCTV on HSI inpainting are reported in the supplementary file.

Although the proposed CTV regularization utilizes the nuclear norm to embed the correlation between the gradient maps, it is still not sufficiently accurate enough. There also remain other priors can be further explored. Thus, it could be interesting to utilize other techniques such as deep image priors to characterize more priors of the gradient maps. Besides, many studies have been devoted to the tensor decompositions, which is expected to more faithfully deliver data intrinsic structures. In fact, if we define the correlated total variation on gradient map of tensor via tensor nuclear norm \cite{lu2016tensor, lu2019tensor}, our method can be readily generalized to tensor data. To better illustrate this, we add the additional experiments and discussions on extending our CTV from matrix form to the tensor form in the supplementary materials. Since the tensor algebra is more complicated than that of matrix, we need to devise more theoretical tools for tensor analysis to answer whether the exact decomposition conditions presented in this work are still hold on these higher-order data. More precise tight upper bound for representing the optimal rank of the recoverable matrix, especially more elaborate theoretical expression on $\mu$, for the 3DCTV-RPCA model is also worthy to be investigated in our future research.

%
%

\ifCLASSOPTIONcaptionsoff
  \newpage
\fi


\bibliographystyle{IEEEtran}
\bibliography{mybibfile}

\ifCLASSOPTIONcaptionsoff
  \newpage
\fi

\end{document}